\documentclass[lettersize,journal]{IEEEtran}
\usepackage{amsmath,amsfonts}
\usepackage{algorithmic}
\usepackage{algorithm}
\usepackage{array}
\usepackage{textcomp}
\usepackage{stfloats}
\usepackage[utf8]{inputenc}
\usepackage{xspace}
\usepackage{url}
\usepackage{blindtext}
\usepackage{hyperref}
\hypersetup{
    colorlinks=true,
    linkcolor=black,
    citecolor=black,
    filecolor=black,
    urlcolor=black,
}
\usepackage{verbatim}
\usepackage{graphicx}
\usepackage[font=footnotesize]{caption,subcaption}  %
\usepackage{tikz, tikzscale}
\usepackage{standalone}
\usepackage{etoolbox}
\usepackage{multirow, makecell, booktabs}
\usepackage{placeins}
\usepackage[absolute,overlay]{textpos}
\usetikzlibrary{
    arrows.meta,
    shapes,
    positioning,
    fit,
    plotmarks,
    fadings,
    patterns
}

\usepackage[inkscapepath=out/svg-inkscape, notransparent]{svg}
\usepackage{pgfplots}
\usepackage{pgfplotstable}
\usepackage{pgfmath}
\usepackage{xfrac}   %
\usepackage{amsthm}

\newtheorem{lemma}{Lemma}

\usepgfplotslibrary{
    colorbrewer,
    groupplots,
    fillbetween,
}

\definecolor{blue}{rgb}{0, 0.5, 0.8}
\definecolor{paleblue}{rgb}{0.5, 0.7, 1}
\definecolor{red}{rgb}{0.82, 0.1, 0.26}
\definecolor{green}{rgb}{0, 0.5, 0.0}
\definecolor{orange}{rgb}{1, 0.7, 0.0}
\definecolor{lightorange}{rgb}{1, 0.8, 0.7}
\definecolor{gray}{rgb}{0.4, 0.4, 0.4}
\definecolor{lightgray}{HTML}{b3b3b3}
\definecolor{trueBlue}{HTML}{0000ff}
\definecolor{myRed}{HTML}{f1615c}    
\definecolor{myBlue}{HTML}{5eb8e7}   
\definecolor{myGreen}{HTML}{a0d186}  
\definecolor{myPurple}{HTML}{776eb2} 

\newcommand{\continuousline}[1]{
    \raisebox{2pt}{
        \tikz{
            \draw[-,#1,solid,line width = 0.9pt](0,0) -- (3mm,0);
        }
    }
}
\robustify{\continuousline}

\usepackage{cite}
\usepackage{flushend}
\usepackage[all]{nowidow}
\usepackage[english]{babel}
\usepackage{siunitx}
\robustify\bfseries
\usepackage{todonotes}
\usepackage[normalem]{ulem}
\newcommand{\redsout}{\bgroup\markoverwith{\textcolor{red}{\rule[0.5ex]{2pt}{0.5pt}}}\ULon}

\usepackage{mathtools}
\usepackage{xspace}
\usepackage{tensor}
\usepackage{mathrsfs}

\newcommand{\func}[1]{\uppercase{#1}}
\newcommand{\funcVector}[1]{\mathbf{#1}}

\newcommand{\Vector}[1]{\mathbf{#1}}
\newcommand{\VectorSym}[1]{\boldsymbol{#1}}
\newcommand{\Matrix}[1]{\boldsymbol{#1}}

\newcommand{\set}[1]{\mathcal{#1}}

\newcommand{\qued}[1]{\blacksquare} %

\newcommand{\estimator}[1]{\hat{#1}}
\newcommand{\optimum}[1]{{#1}^{\ast}}

\newcommand{\minus}{\scalebox{0.6}{$-$}}
\newcommand{\plus}{\scalebox{0.6}{$+$}}
\newcommand{\upperBound}[1]{{#1}^{\plus}}
\newcommand{\lowerBound}[1]{{#1}^{\minus}}
\makeatletter
\DeclareRobustCommand\onedot{\futurelet\@let@token\@onedot}
\def\@onedot{\ifx\@let@token.\else.\null\fi\xspace}

\DeclareRobustCommand\oneperiod{\futurelet\@let@token\@oneperiod}
\def\@oneperiod{\ifx\@let@token,\else,\null\fi\xspace}

\def\eg{e.g.\oneperiod}

\def\ie{i.e.\oneperiod}

\makeatother

\DeclarePairedDelimiter\norm{\lVert}{\rVert}
\DeclarePairedDelimiter\abs{\lvert}{\rvert}

\DeclarePairedDelimiterX{\infdivx}[2]{(}{)}{%
    #1\;\delimsize\|\;#2%
}

\newcommand{\realNumbers}{\mathbb{R}}
\newcommand{\naturalNumbers}{\mathbb{N}}

\newcommand*{\boldone}{\text{\usefont{U}{bbold}{m}{n}1}}

\newcommand{\transpose}[1]{{#1}^{{\mkern-1.0mu}\mathsf{T}}}  %

\newcommand{\covarianceMatrix}{\Matrix{\Sigma}}

\newcommand{\Time}{{t}}

\newcommand{\horizon}{{\Time_{\text{h}}}}
\newcommand{\samplingInterval}{\Time_\text{s}} %
\newcommand{\samplingCount}{n} %

\newcommand{\computationTime}{{\Time_\text{c}}}
\newcommand{\pinnedTime}{{\Time_\text{pin}}}
\newcommand{\pinnedIndices}{{k}}
\newcommand{\weight}{{w}}
\newcommand{\error}{{e}}

\newcommand{\probabilityOperator}{\operatorname{\mathbb{P}}}

\newcommand{\normalDistribution}{{\mathcal{N}}}

\newcommand{\errorFunction}{{\operatorname{erf}}}
\newcommand{\standardDeviation}{{\sigma}}
\newcommand{\meanValue}{\VectorSym{\mu}}

\newcommand{\equalityConstraints}{\Vector{h}}
\newcommand{\inequalityConstraint}{G}
\newcommand{\inequalityConstraints}{\Vector{g}}
\newcommand{\parameterVector}{\VectorSym{p}}
\newcommand{\controlInput}{{\VectorSym{a}}}  %
\newcommand{\costFunctional}{{J}}
\newcommand{\physicalState}{\VectorSym{x}}

\newcommand{\costScalingFactor}{{\weight}}

\newcommand{\cartesianCoordinateFrame}{\mathcal{C}}
\newcommand{\frenetCoordinateFrame}{\mathcal{F}}
\newcommand{\coordinateTransformation}[2]{\tensor*[_{#1}^{#2}]{M}{}}  %
\newcommand{\cartesianFrameX}{\mathrm{x}}
\newcommand{\cartesianFrameY}{\mathrm{y}}

\newcommand{\positionCartesianX}{{x}}
\newcommand{\positionCartesianY}{{y}}

\newcommand{\positionFrenetLon}{{s}}
\newcommand{\positionFrenetLat}{{d}}

\newcommand{\speed}{{v}}

\newcommand{\speedMax}{{\speed}_\text{max}}

\newcommand{\acceleration}{{a}}  %
\newcommand{\deceleration}{\acceleration}

\newcommand{\decelerationComfortMax}{{\acceleration}_{\text{cft}}}
\newcommand{\jerk}{{j}}

\newcommand{\steeringAngle}{{\delta}}

\newcommand{\yawAngle}{{\psi}}

\newcommand{\world}{\ensuremath{\mathcal{W}}}
\newcommand{\environment}{\ensuremath{\mathcal{E}}}

\newcommand{\scene}{\ensuremath{\mathcal{S}}}

\newcommand{\object}{\VectorSym{o}}
\newcommand{\objects}{{\set{O}}}

\newcommand{\timeHeadway}{\Time_\text{hw}}
\newcommand{\risk}{\ensuremath{\alpha}}
\newcommand{\brakingDistance}{\positionFrenetLon_\text{brake}}

\usepackage{balance}
\usepackage[noabbrev, nameinlink]{cleveref}

\begin{document}

\title{Decision-theoretic MPC: Motion Planning with Weighted Maneuver Preferences Under Uncertainty}

\author{Ömer~Şahin~Taş,~\IEEEmembership{Member,~IEEE},
    Philipp Heinrich Brusius, %
    and Christoph~Stiller,~\IEEEmembership{Fellow,~IEEE}
    \thanks{Ömer~Şahin~Taş and Christoph~Stiller are with FZI Research Center for Information Technology and Karlsruhe Institute of Technology (KIT), both 76131 Karlsruhe, Germany. Philipp Heinrich Brusius is with Porsche Motorsport, \mbox{Dr.\ Ing.\ h.c.\ F.\ Porsche AG}, Ostfildern 73744, Germany. (Corresponding author: Ömer~Şahin~Taş, e-mail: \texttt{\url{tas@fzi.de}}).}
}


\markboth{Preprint}%
{Shell \MakeLowercase{\textit{et al.}}: A Sample Article Using IEEEtran.cls for IEEE Journals}


\maketitle
\begin{abstract}
    Continuous optimization based motion planners require specifying a maneuver class before calculating the optimal trajectory for that class.
    In traffic, the intentions of other participants are often unclear, presenting multiple maneuver options for the autonomous vehicle.
    This uncertainty can make it difficult for the vehicle to decide on the best option.
    This work introduces a continuous optimization based motion planner that combines multiple maneuvers by weighting the trajectory of each maneuver according to the vehicle's preferences.
    In this way, the planner eliminates the need for committing to a single maneuver.
    To maintain safety despite this increased complexity, the planner considers uncertainties ranging from perception to prediction, while ensuring the feasibility of a chance-constrained emergency maneuver.
    Evaluations in both driving experiments and simulation studies show enhanced interaction capabilities and comfort levels compared to conventional planners, which consider only a single maneuver.
\end{abstract}

\begin{IEEEkeywords}
    Planning under uncertainty, interaction-aware planning, contingency planning, scenario MPC, safety.
\end{IEEEkeywords}

\section{Introduction}
\label{sec:intro}

\IEEEPARstart{A}{utonomous} vehicles navigate under uncertain environment information, originating from noisy sensor data and limited field-of-view to unknown future actions of other traffic participants.
Despite these uncertainties, they must adhere to their driving preferences and plan a course of motion that is safe and comfortable.
This presents a considerable challenge, as it requires quantifying the current risk and responding to unforeseen developments in the current scenario.
Thus, it is necessary to evaluate the uncertainty, the available maneuver options of other vehicles, as well as the motion limits of the vehicle in the planning process.
While these aspects are recognized within the framework of continuous optimization based receding horizon planning, effectively integrating them into a unified planning scheme remains an ongoing challenge.

\begin{figure}[t!]
    \input{1_figures/phantom/figure_1_final_full.tex}
    \caption{
        A situation where the existence of the red vehicle is unclear to the blue vehicle.
        Conventional planners choose a single maneuver class and solve the optimization problem.
        This results in a motion profile denoted with either $\mathrm{\textcolor{blue}{A}}$ or $\mathrm{\textcolor{blue}{B}}$ in the path-time diagram, as depicted on the left-hand side.
        However, when current information is limited, the vehicle may not be able to determine the optimal maneuver class and choose a defensive one instead.
        Decision-theoretic MPC considers maneuver preferences and yields profile \textcolor{orange}{$\mathrm{C}$}, which stays within the path range of $\mathrm{\textcolor{blue}{A}}$ and $\mathrm{\textcolor{blue}{B}}$ in the hatched time interval on the path-time diagram.
        After $2t_\text{pin}$, it continues with $\mathrm{\textcolor{orange}{C}\textcolor{myPurple}{A}}$ and $\mathrm{\textcolor{orange}{C}\textcolor{myPurple}{B}}$, which will be recalculated in the next planning instance.
        \mbox{Profile $\mathrm{\textcolor{gray}{Z}}$} is the emergency maneuver of \mbox{$\mathrm{\textcolor{orange}{C}}$}.
        The right-hand side depicts corresponding maneuvers of profiles $\mathrm{\textcolor{blue}{A}}$, $\mathrm{\textcolor{blue}{B}}$ and \textcolor{orange}{$\mathrm{C}$} at $2t_\text{pin}$.
    }
    \label{fig:phantom_object_treatment}
\end{figure}

In this work we introduce a model predictive control (MPC) based motion planning framework that addresses all the aforementioned uncertainties.
Unlike existing, \mbox{\textit{conventional}}, continuous optimization based motion planners, our framework does not constrain the planned motion profile to a single maneuver class.
Instead, it incorporates the preferences on multiple maneuvers by including their variables in its optimization parameters.
The resulting motion profile is shaped by weights reflecting these preferences (see~\cref{fig:phantom_object_treatment}).
When there is no clear maneuver preference, this formulation defers maneuver decisions to later times, facilitating interactive behavior and smooth transitions between maneuver options.
In the worst case, the planner applies an emergency maneuver.
Unlike existing planning approaches, it computes the uncertainty in its emergency maneuver for the current state uncertainty, and ensures the feasibility of it with chance-constraints.
We employ gradient-based optimization to solve the resulting nonlinear optimization problem in real-time.

The contributions of this work are as follows:
\begin{itemize}
    \item Motion planning with continuous optimization that is not constrained to a single maneuver class.
          The presented formulation integrates discrete maneuver decisions with continuous optimization variables, enabling smooth transitions between maneuvers or deferring decisions when information is incomplete.
    \item Propagation of vehicle state uncertainties to the emergency maneuver, and ensuring its chance-constrained safety.
          This complements existing state-dependent collision penalties, enabling proactive yet comfortable navigation under limited field-of-view and in the presence of rule-violating traffic participants.
    \item Evaluation of warm-start solver performance across vehicle models.
          Considering our multiple objectives and cost terms, we analyze the impact of vehicle model complexity on replanning times.
    \item Demonstration of real-world effectiveness through vehicle tests in intersection crossing scenarios. The effectiveness is further validated by simulations leveraging real-world data, and by comparisons with a conventional MPC planner in a separate scenario.
\end{itemize}

The remainder of this paper is structured as follows:
\cref{sec:related_work} presents related work and identifies open problems in motion planning under uncertainty.
Following this, \cref{sec:environment} introduces our environment model, outlining the assumptions employed therein, which subsequently serve as the foundation for defining safety constraints discussed in \cref{sec:safety}.
\Cref{sec:method} frames motion planning as an optimization problem and outlines the scope of applied constraints.
Given the complexity of solving the proposed optimization problem, \cref{sec:optimization} explores solution approaches.
The evaluations of our planning approach are then demonstrated in \cref{sec:evaluation}, followed by discussions in \cref{sec:discussions}.
The paper concludes with \cref{sec:conclusion}, highlighting the main contributions and providing directions for future research.

\section{Related Work}
\label{sec:related_work}

Motion planning algorithms should focus on five key objectives: i) arriving at the \textit{destination}, ii) ensuring \textit{comfort}, iii) minimizing \textit{risk}, iv) facilitating \textit{interaction}, and v) conducting \textit{information gathering} \cite{tas2022motion}.
Achieving these objectives require accounting for uncertainties, including i) the uncertainty in \textit{physical state}, which typically covers the uncertainty in position and velocity,  ii) the uncertainty in \textit{prediction}, which encompasses the uncertainties in future route choice and intentions, and iii) the uncertainty in \textit{existence}, which reflects the uncertainty of an object to exist, either inside or beyond the \mbox{field-of-view \cite{gonzalez2015review,paden2016survey,sana2023autonomous, tas2020tackling}}. 

Forecasting the future motion of other traffic participants in interactive driving scenarios involves considering multiple maneuver options that must be coherent for all of the agents \cite{lefevre2014survey, brown2020modeling}.
This poses a substantial challenge due to the difficulty of resolving a scene-wide set of motion sequences that all agents follow.
As a result, forecasting algorithms often cannot settle on a single scene-wide mode, resulting in uncertainty about the most likely options.
While it is already difficult to forecast states for a single scene-wide mode, multiple hypotheses expand the range of possible trajectories further.  %

Each scene-wide mode corresponds to a maneuver class for the autonomous vehicle to consider.
The trajectories within a single maneuver class are defined by a cost function reflecting the driving preferences \cite{kochenderfer2022decision}[p.\ 112], and are homotopic, meaning they can be continuously deformed into one another while maintaining their start and end points \cite{bender2015combinatorial,altche2017partitioning,patterson2021optimal}.
Planning for multiple maneuvers therefore requires discrete decisions typically represented with integer variables or \textit{logic constraints} \cite{tas2014integrating,park2015homotopy, qian2016optimal, burger2018cooperative}.
However, under uncertainty, as mentioned in the preceding section, forecasting algorithms often yield maneuver probabilities rather than definitive decisions.
Consequently, motion planners must evaluate multiple maneuvers simultaneously with their corresponding likelihood.
Planning with multiple maneuver options without rendering a defensive behavior requires in-depth analysis of existing approaches for safety checks.
Therefore, we first inspect existing approaches for performing safety checks, and then focus on motion planning under uncertainty.
Specifically, we inspect local-continous optimization based methods, as well as approaches that do not require predefined maneuver classes and approaches that leverage information gathering.

\subsection{Approaches for Safety Checks Under Uncertainty}

Various approaches exist to ensure safety under uncertainty, each with its own set of strengths and limitations.
Early work relied on Time-To-X metrics, where ``X'' represents a selected event such as collision (TTC), the last possible reaction (TTR), or braking (TTB), all measured along the longitudinal dimension \cite{minderhoud2001extended,hillenbrand2006efficient, zhang2006new, hillenbrand2007fahrerassistenz}.
Later studies aimed to generalize these metrics into Cartesian space \cite{ward2014vehicle} and model them as probability distributions \cite{eidehall2008statistical, berthelot2011handling, berthelot2012novel, stellet2015uncertainty, stellet2016analytical}.
However, these metrics do not scale well for longer time horizons \cite{medina2023speed}.


A common approach to ensure safety is preventing any overlap between an agent's occupancy and obstacles.
While these occupancies can be represented with polygons \cite{schulman2014motion}, circles are often preferred as they enable differentiable overlap calculations \cite{ziegler2010fast,ziegler2014trajectory}.
The use of circle-based shapes is primarily employed in robust approaches, such as \textit{formal verification}, as they do not inherently correspond to parametric probability distributions.
Variations of these methods, as used in autonomous driving, apply set-based over-approximations to model the reachable states of traffic participants over a given prediction horizon \cite{althoff2007online,althoff2007safety}
A major disadvantage of reachable set-based verification is that over-approximations can quickly grow over time and cause defensive behavior.


An alternative approach involves using probability distributions to represent uncertainty.
Considering that uncertainties are often modeled as Gaussian distributions, using ellipses to represent positions becomes a more suitable choice, enabling a direct association with confidence levels \cite{brito2019model}.
Consequently, this connection allows for closed-form solutions \cite{calafiore2006distributionally, blackmore2006probabilistic} when safety is defined as a chance-constraint with a prescribed risk level \cite{charnes1963deterministic}.
There are alternative ways to constrain the risk, such as Value-at-Risk (VaR) or conditional-VaR \cite{majumdar2020should}.
Although these constraints limit the likelihood of an outcome, they do not exclude its occurrence.

Interaction capabilities between traffic participants are often overlooked while evaluating safety.
\textit{Responsibility-Sensitive-Safety} (RSS) addresses this gap and formalizes traffic rules through mathematical models.
It ensures safety as long as the vehicles' motion stays within a set range of parameterized limits \cite{shalevshwartz2017formal}.
While recent work has enhanced RSS to overcome its conservative results \cite{sidorenko2022towards} or to account for uncertainties \cite{nyberg2021risk}, RSS still has a significant drawback.
Its reliance on 40 parameters complicates its generalizability \cite{naumann2021on}.

\subsection{Planning with Local-Continuous Methods}

Under uncertainty, absolute safety cannot be guaranteed.
Consequently, existing approaches aim to minimize risk.
A different line of research uses numerical optimization to solve the planning problem and ensures safety via stochastic control techniques \cite{vitus2011feedback, vitus2013probabilistic, carvalho2014stochastic, lenz2015stochastic, de2014collision, lefkopoulos2021trajectory}.
These methods handle uncertainties in prediction in the same way as they handle state uncertainties.
To remain viable for real-time applications, these methods approximate uncertainties with parametric probability distributions and apply \emph{chance-constraints} to minimize collision risk.

Several works leverage formal verification to ensure the existence of a collision-free maneuver until the next planning instance \cite{petti2005safe}.
If the next planning instance cannot ensure safety, the system executes the previously planned emergency maneuver.
This strategy is often referred to as \emph{fail-safe} motion planning \cite{magdici2016fail}.
Reachability analysis is applied to guarantee the availability of emergency maneuvers against prediction uncertainty and occlusions \cite{orzechowski2018tackling, debada2020occlusion}.

To address such \emph{scenario} uncertainties, a more effective approach is to couple optimal motion planning with fail-safe motion plans.
This is achieved by integrating the emergency maneuver as a constraint in the numerical optimization problem \cite{tas2018limited}.
In contrast to fail-safe motion planning, which calculates emergency maneuvers as a second step after optimizing the desired trajectory \cite{magdici2016fail}, this approach modifies the optimal trajectory to ensure the feasibility of the emergency plan.
To distinguish this approach from fail-safe planning, we call it a \textit{fallback} maneuver.
Throughout the rest of this work, we use this term to describe emergency maneuvers integrated into optimal motion planning.

The work presented in \cite{tas2018limited} covers driving scenarios with uncertain position and speed measurements, and occlusions.
The feasibility of an emergency braking maneuver is ensured amidst these uncertainties by using chance-constraints.
Another work takes a similar approach, but instead of addressing uncertainties, it decouples longitudinal and lateral planning to use lateral swerve maneuvers as a fallback \cite{pek2018failsafe}.
A subsequent work extends this method by considering measurement uncertainties and ensuring safety through chance-constraints \cite{brudigam2021stochastic}.

Another line of work plans multiple maneuvers simultaneously.
These approaches analyze the scene-wide modes and create \textit{contingency} trajectories for each scenario.
While the cost function can be identical for all maneuvers, it is also possible to use a separate cost function for the maneuvers.
For instance, one work applies this approach to fallback maneuvers on slippery roads \cite{alsterda2019contingency}.
Another similar approach is to perform the decision \emph{branching} at arbitrary times in the planning horizon \cite{chen2022interactive, oliveira2023interaction, gaetan2024scenario}.
While these approaches cover uncertainties within a single maneuver class and plan trajectories accordingly, they do not address the probability of choosing one scenario or branch over the others.

\subsection{Planning Without Predefined Maneuver Classes}

Some planning methods do not require a predefined maneuver class for planning.
Sample-based search methodologies, such as rapidly-exploring random trees \cite{luders2010chance, aoude2013probabilistically}, and lattice-based search algorithms \cite{artunedo2020motion} fall into this category.

While these planning methods primarily concentrate on state uncertainties and apply probability constraints,
a more general approach is to frame motion planning under uncertainty as a partially observable Markov decision process (POMDP).
The POMDP framework allows for a seamless integration of state, prediction \cite{ulbrich2013probabilistic, brechtel2014probabilistic, hubmannAutomatedDrivingUncertain2018}, and existence uncertainties \cite{boutonScalableDecisionMaking2018, hubmann2019pomdp, wray2021pomdps}.
Furthermore, like other sampling-based methods, the planned motion is not necessarily bound to maneuver classes.
Despite efforts to improving sampling efficiency \cite{tas2020efficient}, these methods are still outperformed by planning approaches using gradient-based optimization in terms of runtime efficiency.

\begin{figure}
    \centering
    \fontsize{8}{9}\selectfont
    \def\svgwidth{1.0\columnwidth}
    \scalebox{1.0}{\includesvg{1_figures/environment/plot_distance_3}}
    \caption{
        An intersection crossing scene with the ego vehicle depicted in blue.
        Nearest conflict points of other vehicles along the ego vehicle's path are marked on a vertical axis to the right with $s_1$ and $s_2$.
        Considering the red vehicle's two potential routes, the point on the route that is nearest to the ego vehicle's path marks the start of the overlap.
        While the overlap with the red vehicle is limited, the overlap with the green vehicle could potentially continue indefinitely, as it follows the ego vehicle's route.
    }
    \label{fig:plot_distance}
\end{figure}

\subsection{Information Gathering Planning and Others}

Several works aim to reduce the current uncertainty about the environment, and maximize the information.
Some works integrate occlusion inference into planning and rationalize potential traffic participants \cite{yu2019occlusion, wang2021reasoning, sanchez2022foresee, zhao2024inference}.
Others define the maximization of visible field as an additional driving objective and perform lateral offsetting to increase it \cite{andersen2019trajectory}.
A promising approach is the execution of dedicated actions to actively maximize the vehicle's information of other vehicles' internal states, such as driving intentions \cite{sadigh2016information}, \cite[p.\ 127]{tas2022motion}.

Lastly, there is a growing body of work using deep neural networks to learn the structure of the problem.
The approaches that offer safety guarantees are built on the components of existing planning works:
they either select a motion plan from a set of safe plans \cite{zeng2019end, casas2021mp3} or employ the aforementioned safety check methods to maintain safety \mbox{under uncertainty \cite{kamran2021minimizing}}.

%

\section{Environment Model}
\label{sec:environment}

An autonomous vehicle operates within a subset of the \textit{world} $\world \subseteq \world_\forall$ that contains a set of objects \mbox{$\objects = \{ \object_0, \ldots, \object_{k_\world} \}$}, where $k_\world \in \naturalNumbers_0$.
An object $\object_i$ can belong to a variety of classes, such as cars, pedestrians, cyclists, or even traffic lights.
The \textit{state} $\physicalState^{(\Time)}_i$ of an object $i$ stores the relevant parameters necessary to describe the system at time $\Time \in \naturalNumbers$.
Sensors of an object $i$ can perceive only a limited portion of the world $\world$, which we refer to as the \textit{environment} $\environment$.
The environment contains a subset of the objects in its surroundings $\objects_{\environment} = \{ \object_i {\}}^{k_{\environment}}_{i=1}$, where ${k_{\environment} \leq k_\world}, k_{\environment} \in \naturalNumbers$, and $i = 0$ is reserved for the ego vehicle.
The environment may include objects that were not detected previously.

Sensor data is typically imperfect, and therefore, state measurements $\estimator{\VectorSym{x}}$ carry a degree of uncertainty.
These measurements are often modeled with additive noise such that $\estimator{\VectorSym{x}}  = \VectorSym{x} + \VectorSym{\zeta}_{\VectorSym{x}}$, where $\VectorSym{\zeta}_{\VectorSym{x}}$ follows a Gaussian distribution with a zero-mean vector and a positive definite covariance matrix, \ie \mbox{$\VectorSym{\zeta}_{\VectorSym{x}}  \sim \normalDistribution \bigl( \VectorSym{0}, \covarianceMatrix_{\VectorSym{x}} \bigr)$}.

Traffic rules are often defined along the lanes of a route.
Sections of lanes with consistent rules and topology are known as \textit{lanelets} \cite{bender2014lanelets}.
Lanelet maps contain essential information about potential conflict areas and route centerlines \cite{poggenhans2020pathfinding}, as depicted in \cref{fig:plot_distance}.
The centerlines are modeled by polylines and are used to define a \textit{Frenet-Serret frame} along the route of the vehicle.
This facilitates the identification of driving intentions of other vehicles.
We denote the transformation from Cartesian coordinates to Frenet-coordinates as
\begin{equation} \label{eq:cartesian2frenet}
    {(
        {\positionFrenetLon},
        {\positionFrenetLat}
        )}
    =
    \coordinateTransformation{\cartesianCoordinateFrame}{\frenetCoordinateFrame} (
    \positionCartesianX,
    \positionCartesianY
    ) \, ,
\end{equation}
and back
\begin{equation} \label{eq:frenet2cartesian}
    {(
        {\positionCartesianX},
        {\positionCartesianY}
        )}
    =
    \coordinateTransformation{\frenetCoordinateFrame}{\cartesianCoordinateFrame} (
    \positionFrenetLon,
    \positionFrenetLat
    ).
\end{equation}
It is important to note that this transformation is not always bijective \cite{tas2014integrating}.
We define agents that influence decision-making as scene objects and denote by $\objects_{\scene}$, where $\objects_{\scene} \subseteq \objects_{\environment}$.
For instance, in \cref{fig:plot_distance}, the red, green, and black vehicles are considered scene objects, whereas the beige vehicle is not.

\section{Maintaining Safety Under Uncertainty}
\label{sec:safety}

It is essential to address uncertainties without resorting to defensive behaviors.
Previous work has shown that fallback plans can provide a safety net against the worst-case scenarios.
To make the fallback plan robust to uncertain state measurements, we first derive the uncertainty in the fallback plan and leverage chance-constraints to minimize the collision risk under the worst-case event.
Then we extend the derived formulations to handle both existence and prediction uncertainties.

\subsection{Fallback Plans Under State Uncertainty}

The fallback maneuver should be both simple to compute and consistently reduce collision risk when executed.
Given these requirements, and considering that traffic rules and maneuver intentions are analyzed using longitudinal coordinates along roads, we choose full braking as the fallback plan.
We refer to this full braking maneuver as the \textit{Z-plan} (see \cref{fig:phantom_object_treatment}), signifying the vehicle's last available option -- ceasing movement.

Using full braking as fallback has computational advantages when state uncertainties are modeled with Gaussian distributions.
A linear transformation of a Gaussian distributed variable $\VectorSym{z} \sim \normalDistribution(\meanValue, \covarianceMatrix)$ results in $\VectorSym{z}_T \sim \normalDistribution (\Matrix{R} \meanValue + \VectorSym{t}, \Matrix{R} \covarianceMatrix \transpose{\Matrix{R}})$, where $\Matrix{R}$ is the rotation matrix and $\VectorSym{t}$ is the translation vector.
Projecting a zero-mean bivariate Gaussian distribution onto a segment of centerline via a unit length vector $\VectorSym{u} \in \realNumbers^{2}$ results in a zero mean univariate Gaussian with variance
\begin{equation}
    \label{eq:projection}
    \sigma^2_{\VectorSym{u}} = \lambda_{\VectorSym{u}} (\covarianceMatrix) = {\VectorSym{u}} \covarianceMatrix \transpose{\VectorSym{u}}.
\end{equation}

Modeling measurement uncertainties with a bivariate Gaussian distribution and projecting it onto a polyline results in a univariate Gaussian distribution of noise.
Specifically, we model position, velocity, and acceleration measurements with independent, additive, zero-mean Gaussian noise.
Projecting these measurements onto the route centerline gives
$
    \estimator{\positionFrenetLon} = \positionFrenetLon + \zeta_\positionFrenetLon \;\; \text{with} \;\; \zeta_\positionFrenetLon \sim \normalDistribution \bigl( 0, {\standardDeviation_\positionFrenetLon^2} \bigr),
$
$
    \estimator{\speed} = \speed + \zeta_\speed \;\; \text{with} \;\; \zeta_\speed \sim \normalDistribution \bigl( 0, {\standardDeviation_\speed^2} \bigr),
$
and
$
    \estimator{\acceleration} = \acceleration + \zeta_\acceleration \;\; \text{with} \;\; \zeta_\acceleration \sim \normalDistribution \bigl( 0, {\standardDeviation_\acceleration^2} \bigr),
$
respectively.

\begin{lemma}
    The first-order Taylor expansion of the braking distance function with respect to velocity and deceleration, under the assumption of independent $\zeta_\positionFrenetLon$ and $\zeta_\speed$, yields a braking distance estimate with an additive, zero-mean Gaussian noise.
\end{lemma}

\begin{proof}
    The first-order Taylor expansion of a function $F (\VectorSym{z})$ at $\VectorSym{z}^{(0)}$ is given as
    \begin{equation}
        F (\VectorSym{z})
        \approx
        F \bigl( \VectorSym{z}^{(0)} \bigr) +
        \nabla F \big\rvert_{\VectorSym{z}^{(0)}}
        \bigl(
        \VectorSym{z} - \VectorSym{z}^{(0)}
        \bigr).
    \end{equation}
    As $\VectorSym{z}^{(0)} = 0$ for additive Gaussian noise, the variance is obtained as
    \begin{equation} \label{eq:linearied_error}
        \standardDeviation^2_{F} \approx
        {
            \left(
            \frac{\partial F}{\partial \VectorSym{z}_0}
            \right)
        }^2
        \sigma^2_{\VectorSym{z}_0}
        +
        \ldots
        +
        {
        \left(
        \frac{\partial F}{\partial \VectorSym{z}_{n-1}}
        \right)
        }^2
        \sigma^2_{\VectorSym{z}_{n-1}}.
    \end{equation}
    Applying this on braking distance
    \begin{equation}
        \label{eq:braking_dist}
        \brakingDistance = F_\text{brake} (\speed , \deceleration) = \frac{\speed^2}{2\deceleration} \, ,
    \end{equation}
    yields
    \begin{align}
        \standardDeviation^2_{\text{brake}}
         & \approx
        {
            \left(
            \frac{\partial F_{\text{brake}} }{\partial \speed}
            \right)
        }^2
        \sigma^2_{\speed}
        +
        {
        \left(
        \frac{\partial F_{\text{brake}} }{\partial \deceleration}
        \right)
        }^2
        \sigma^2_{\deceleration} \nonumber \\
         & \approx
        {
            \biggl(
            \frac{\speed}{\deceleration}
            \biggr)
        }^2
        \sigma^2_{\speed}
        +
        {
        \left(
        -\frac{ {\speed}^2 }{ 2 {\deceleration}^2 }
        \right)
        }^2
        \sigma^2_{\deceleration} \, ,
    \end{align}
    which corresponds to
    \begin{equation} \label{eq:braking_distance_normal}
        \estimator{\positionFrenetLon}_\text{brake} = \brakingDistance + \zeta_\text{brake} \;\; \text{where} \;\;\zeta_\text{brake} \sim \normalDistribution (0, \sigma^2_{\text{brake}}).
    \end{equation}
\end{proof}

\begin{lemma}
    The stop distance can be modeled with an additive white Gaussian noise if the current position and the braking distance are independently modeled with their respective zero-mean additive white Gaussian noises.
\end{lemma}

\begin{proof}
    \begin{align}
        \estimator{\positionFrenetLon}_\text{stop} = F_{\text{stop}} (\positionFrenetLon, \speed, \deceleration)
         & = \estimator{\positionFrenetLon} + {\estimator{\positionFrenetLon}}_{\text{brake}} \nonumber                                                  \\
         & = \positionFrenetLon + \zeta_{\positionFrenetLon} + \brakingDistance + \zeta_\text{brake}                                           \nonumber \\
         & = {\positionFrenetLon}_\text{stop} + \zeta_\text{stop} \, ,                                                                                   %
    \end{align}
    with ${\positionFrenetLon}_\text{stop} = \positionFrenetLon + \brakingDistance$ and $\zeta_\text{stop} \sim \normalDistribution \bigl(0, \sigma^2_{\text{stop}} \bigr)$, where $\standardDeviation^2_\text{stop} = \standardDeviation^2_\speed + \standardDeviation^2_{\text{brake}}$.
\end{proof}

\subsection{Chance-Constraints on Gaussian Distributions}

We constrain the distribution of the stop position along the centerline using chance-constraints.
For a vector of random variables $\VectorSym{z}$, we can constrain the probability that the absolute deviation of a function $\func{f}$ from its expected value does not exceed a certain threshold $\theta$
\begin{equation} \label{eq:chance_constraints_general}
    \probabilityOperator
    \Bigl(
    \big\vert
    \func{f} (\VectorSym{z} + \VectorSym{\zeta}_{\VectorSym{z}})
    -
    \mathbb{E} \bigl(  \func{f} (\VectorSym{z} + \VectorSym{\zeta}_{\VectorSym{z}}) \bigr)
    \big\vert
    \leq
    \theta
    \Bigr)
    \geq
    1 - \risk \, ,
\end{equation}
where $\risk$ represents a predefined numerical value, often referred to as \textit{risk}.
For zero-mean Gaussian noise, this can be expressed as
\begin{equation}
    \probabilityOperator (\VectorSym{u}_i \estimator{\VectorSym{x}}_i - \VectorSym{\theta}_i \leq 0) \geq 1 - \risk \, ,
\end{equation}
or using \cref{eq:projection} as
\begin{align} \label{eq:cumulative_distribution_func}
    \VectorSym{u}_i  {\VectorSym{x}}_i + \sqrt{ \lambda_{\VectorSym{u}_i} (\covarianceMatrix_i) } \cdot {\phi}^{-1} ( 1 - \risk) \leq \VectorSym{\theta}_i \, , \\
    \intertext{where}
    {\phi}^{-1} (1 - \alpha) = \sqrt{2} {\errorFunction}^{-1} (1 - 2 \alpha) \nonumber.
\end{align}
Computationally efficient approximations of the quantile function ${\phi}^{-1}$ \cite[p.~213]{press1992numerical} allow to use this constraint in \mbox{real-time applications}.

\subsection{Limiting the Collision Risk of Fallback Plans}

Dealing with uncertainties by ensuring the feasibility of the full braking maneuver can promote proactive behavior without compromising safety.
We identify four baseline scenarios for our analysis.

\subsubsection{Follow Drive}

A crucial safety requirement for the ego vehicle $\object_{\mkern 1mu 0}$ is to maintain a safe following distance from the vehicle directly ahead, $\object_k$.
We define the worst-case scenario as the instance when $\object_k$ initiates full braking immediately after $\object_{\mkern 1mu 0}$ processes environment information.
In this scenario, safety is maintained by ensuring that the difference in the stop positions between the two vehicles, $\Delta {\positionFrenetLon}_\text{stop}$, is less than the required standstill distance, $\positionFrenetLon_{\text{min}}$.
The difference in stop positions can be approximated as
\begin{align} \label{eq:stop_position_gaussian_delta_function}
    \Delta {\positionFrenetLon}_{\text{stop}}
     & = F_{\text{stop}, 0}
    \left(
    \positionFrenetLon_0,
    \speed_0,
    \deceleration
    \right)
    -
    F_{\text{stop}, k}
    \left(
    \positionFrenetLon_{k},
    {\speed}_{k},
    {\deceleration}
    \right) \nonumber       \\
     & \approx
    \estimator{\positionFrenetLon}_{\Delta \text{stop}}
        = {\positionFrenetLon}_{\Delta \text{stop}} +
    \zeta_{\Delta \text{stop}} \, ,
\end{align}
where
${\positionFrenetLon}_{\Delta \text{stop}} = {\positionFrenetLon}_{\text{stop}, 0} - {\positionFrenetLon}_{\text{stop}, k}$ and $\zeta_{\Delta \text{stop}}$ is normally distributed as  $\normalDistribution(0, \sigma^2_{\Delta \text{stop}})$, with $\sigma^2_{\Delta \text{stop}} = \sigma^2_{\text{stop}, 0} + \sigma^2_{\text{stop}, k}$.
This implies a chance-constraint of the form
\begin{equation} \label{eq:chance_constraints_braking_delta}
    \probabilityOperator
    \left(
    \estimator\positionFrenetLon_{\Delta {\text{stop}}}
    +\positionFrenetLon_\text{min}
    \leq 0
    \;\lvert\;
    \boldone_{\text{stop}} = 1
    \right)
    \geq 1 - \risk,
\end{equation}
where $\boldone_{\text{stop}}$ is an indicator function that activates the constraint when it is equal to 1, and deactivates it otherwise.
For instance, in the case of an unresolved phantom object (see \cref{fig:phantom_object_treatment}), this constraint must remain active.

\subsubsection{Limited visibility}

When no agent is visible within the field-of-view along the driving corridor (see \cref{fig:limited_visibility_free_driving}), the planner must ensure that it can stop within the visible free distance, $\positionFrenetLon_\text{vis}$, while also maintaining a safety margin, $\positionFrenetLon_\text{min}$.
This requirement is formalized through the chance constraint
\begin{equation} \label{eq:chance_constraints_braking}
    \probabilityOperator
    \left(
    \estimator{\positionFrenetLon}_{\text{stop}}
    - \positionFrenetLon_\text{vis}
    + \positionFrenetLon_\text{min}
    \leq 0
    \; \lvert \;
    \boldone_{\text{vis}} = 1
    \right)
    \geq 1 - \risk.
\end{equation}
\begin{figure}[h!]
    \fontsize{7}{8}\selectfont %
    \vspace{-4mm}
    \centering
    \input{1_figures/free_driving/main.tex}
    \caption{
        A driving scenario with a limited field-of-view.
        The braking distance uncertainty of the blue vehicle is modeled with a Gaussian distribution.
    }
    \label{fig:limited_visibility_free_driving}
\end{figure}

\subsubsection{Yield Intersections}
\label{sec:limited_visibility_at_intersections}

At intersections with obstructed visibility (see \cref{fig:limited_visibility_intersection}), the planner must determine whether it can safely proceed when there might be occluded oncoming vehicles.
To do this, it first calculates the visible distance along the intersecting route, $\positionFrenetLon_{\text{vis}}$.
It then ensures that this visible distance exceeds the required distance, $\positionFrenetLon_{\text{req}}$, which allows an oncoming vehicle to either pass or comfortably decelerate $\decelerationComfortMax$ to the ego vehicle's speed, $\speed_0$, while maintaining a time headway $\timeHeadway$.
The required distance can be calculated as
\begin{align} \label{eq:intersection_required_deceleration}
    \positionFrenetLon_{\text{req}} & =
    S_{\text{req}} (\speed_0 , \speed_k, {\decelerationComfortMax}) = \speed_k \, \Time_{\text{dec}}
    + \frac{1}{2} \, \decelerationComfortMax \, \Time^2_{\text{dec}}
    + \speed_{0} \timeHeadway  \, , \nonumber \\
    \intertext{by substituting $\Time_{\text{dec}} = \sfrac{\speed_k - \speed_{0} }{\abs{\decelerationComfortMax}}$ and $\timeHeadway = \SI{2.0}{\second}$, we obtain}
                                    & =
    \speed_k \, {\left( \frac{\speed_k - \speed_{0} }{{\abs{\decelerationComfortMax}}} \right)}
    + \frac{1}{2} \, {\decelerationComfortMax} \, {\left( \frac{\speed_k - \speed_{0} }{{\abs{\decelerationComfortMax}}} \right)}^2
    + 2 \speed_{0}.
\end{align}

\begin{figure}[h!]
    \fontsize{8}{9}\selectfont
    \centering
    \resizebox{0.8\linewidth}{!} {
        \input{1_figures/intersection_crossing/intersection_crossing.tex}
    }
    \caption{
        An intersection crossing scenario with limited field-of-view.
        The merge point (MP), where the routes intersect, is centrally positioned in both corridors for illustrative purposes.
    }
    \label{fig:limited_visibility_intersection}
\end{figure}

If $\positionFrenetLon_{\text{vis}} < S_{\text{req}} (\speed_{0} , \speedMax, \decelerationComfortMax)$, the planner must satisfy the chance-constraint
\begin{equation} \label{eq:chance_constraints_braking_MP_general}
    \probabilityOperator
    \left(
    \estimator{\positionFrenetLon}_{\text{stop}}
    - \positionFrenetLon_\text{MP}
    + \positionFrenetLon_\text{min}
    \leq 0
    \; \lvert \;
    \boldone_{\text{yield}} = 1
    \right)
    \geq 1 - \risk \, ,
\end{equation}
which may cause the vehicle to slow down.
Otherwise, if $\positionFrenetLon_{\text{vis}} \geq S_{\text{req}} (\speed_{0}, {\speed}_{\text{max}}, \decelerationComfortMax)$, the vehicle can proceed through the intersection, or perform a full stop if a stop sign is present.

\subsubsection{Right-of-Way Intersections}

In intersection scenarios, the intentions of oncoming vehicles may remain unclear, even when considering traffic rules.
While the ego vehicle might have the right-of-way and not bear direct responsibility for a potential accident, it should still consider the possibility of the oncoming vehicle violating traffic rules if its intentions are unclear.
In such cases, the ego vehicle should satisfy the constraint in \cref{eq:chance_constraints_braking}.
Additionally, \cref{eq:intersection_required_deceleration} can be used to assess whether the oncoming vehicle intends to yield.

\section{Decision-Theoretic Model Predictive Control}
\label{sec:method}

Our method finds the optimal action sequence that minimizes an objective function under safety constraints within a receding horizon.
We present the individual components before introducing our planner.

\subsection{Safety in Receding Horizon Planning}

A planning horizon of length $\horizon$ can be approximated using $n$ \textit{colocation points}, also referred to as \textit{trajectory support points}, when sampled with a sampling interval of $\samplingInterval$.
They are denoted by $\parameterVector_i = {[\physicalState_i, \controlInput_{i}]}$ with $0 \leq i \leq \samplingCount$, where $\physicalState$ and $\controlInput$ represent states and actions, respectively.
Not all points can be freely adjusted during optimization, as during computation $\computationTime$, a number of control inputs, $\pinnedIndices \, = \lceil \pinnedTime / \samplingInterval \rceil$ with $\computationTime \, \leq \pinnedTime$, must be held fixed to compensate for time delays and maintain temporal stability.

In the next replanning instance, the first $\pinnedIndices$ parameters from the current instance will be discarded, and the parameters from $\pinnedIndices$ to $2\pinnedIndices$ will be held fixed.
The optimization parameter vector can thus be represented as
\begin{equation} \label{eq:parameters_planning}
    \parameterVector =
    [
    \underbrace{ \parameterVector_{0}, \, \parameterVector_{1:\pinnedIndices} }_{\text{pinned}}, \,
    \underbrace{ \parameterVector_{\pinnedIndices + 1 : 2\pinnedIndices} }_{\text{executed next} }, \,
    \underbrace{ \parameterVector_{2\pinnedIndices + 1 : n } }_{\text{will be recalculated}}
    ]. %
\end{equation}
Though $\parameterVector_{0}$ represents the current measurement, we include it in the parameter vector for the sake of completeness.

Since the actions up to $2\pinnedIndices$ are irreversible, it is crucial for the planner to ensure that none of these actions result in an \textit{inevitable collision state} beyond this interval \cite{petti2005safe,tas2018limited}.
For this purpose, we apply the stop distance chance-constraints, discussed in the previous section, within this time interval.

Chance constrained fallback plans constrain the stop position and enable proactive safety against the worst case scenario.
To ensure collision avoidance beyond just the longitudinal direction, we introduce an additional spatial constraint.
We model the area covered by the ego vehicle with three circles \cite{ziegler2010fast}.
To prevent collisions, these circles must not overlap with the occupancy of any other object $\object$ in the scene, which we model with ellipses.
For each circle centered at $(x, y)$ with radius $r$, we enforce a constraint against an ellipse centered at $(x_{\object}, y_{\object})$ with semi-major and semi-minor axes lengths ${{\sigma}_{\cartesianFrameX, \object}}$ and ${{\sigma}_{\cartesianFrameY, \object}}$
\begin{equation} \label{eq:collision_ellipse}
    \bigwedge_{k = 1}^{3}
    \bigwedge_{\object \in {\objects_{\mkern-2.5mu\scene}}}
    \Bigg[
    \sqrt{
    {\left( \frac{x_{k} - x_{\object}}{ {{\sigma}_{\cartesianFrameX, \object}} + r_k + \varepsilon} \right)}^2
    +
    {\left( \frac{y_{k} - y_{\object}}{ {{\sigma}_{\cartesianFrameY, \object}} + r_k + \varepsilon} \right)}^2
    } > 1 \Bigg] \, ,
\end{equation}
where $\varepsilon$ is an offset that enlarges the ellipse to encapsulate the Minkowski sum of the circle and the ellipse \cite{brito2019model}. %

\subsection{Vehicle Models}
\label{sec:vehicle_model}

The trajectory support points must adhere to the non-holonomic motion constraints of the vehicle.
There are various vehicle models to capture the nonlinear dynamics of the system \mbox{$\dot{\physicalState} = \funcVector{g}_\text{veh} (\physicalState, \controlInput)$} \cite{milliken1995race, polack2017kinematic, goh2020toward}.
In our evaluation, we benchmark several models, including the point mass, the kinematic bicycle, the linear kinematic bicycle, and the dynamic bicycle model with linear tire model.
For the sake of simplicity, we use $\physicalState = {[\positionCartesianX, \positionCartesianY, \yawAngle, \speed]}$ and $\controlInput  = {[\steeringAngle, \acceleration]}$, regardless of the underlying motion model.
We integrate piecewise constant actions with the classic Runge-Kutta method.

\subsection{Driving Objectives and Cost Function}

We define three types of cost term templates for rendering comfortable and safe driving objectives \cite{tas2018making}.
These include \emph{value terms} that penalize deviations from a desired value $\optimum{x}$
\begin{subequations}
    \begin{equation}
        \costFunctional_\text{val} (x) = \weight_\text{val} \, { \norm{\optimum{x} - x} }^2 \, ,
    \end{equation}
    \emph{range terms}, that penalize values outside the interval $[\lowerBound{x}, \upperBound{x}]$
    \begin{equation}
        \costFunctional_\text{ran} (x) =
        \begin{cases}
            \weight_\text{ran} \, {\norm{x - \upperBound{x}}}^2 & \text{if } \upperBound{x} < x                         \\
            0                                                   & \text{if }  \lowerBound{x} \leq x \leq \upperBound{x} \\
            \weight_\text{ran} \, {\norm{\lowerBound{x} - x}}^2 & \text{otherwise,}
        \end{cases}
    \end{equation}
    and \emph{asymmetric terms} around $\optimum{x}$
    \begin{equation}                                                           \\
        \costFunctional_{\text{asym}} (x) =
        \begin{dcases}
            \costScalingFactor_{\text{asym}} \ {\norm{x - \optimum{x}}}^2                              & \text{if } x \geq  \optimum{x} \\
            \costScalingFactor_{\text{asym}} \ \log \! \left( 1 + { \norm{x - \optimum{x}} }^2 \right) & \text{otherwise,}
        \end{dcases}
    \end{equation}
    where values lower than $\optimum{x}$ are penalized with a tolerant loss, such as the Cauchy loss \cite{barronGeneralAdaptiveRobust2017}.
\end{subequations}
It is important to note that both $\costFunctional_\text{ran}$ and $\costFunctional_{\text{asym}}$ include conditional statements.

We calculate the costs over the entire planning horizon.
Therefore, the cost functional $\costFunctional (\parameterVector)$ is defined as the sum of the costs for each element in the vector
\begin{equation}
    \costFunctional (\parameterVector) =
    \sum_{i=0}^n {
    \costFunctional (\VectorSym{\parameterVector}_{i})
    }.
\end{equation}

The total cost for planning consists of
\begin{equation} \label{eq:cost_total}
    \costFunctional_\text{total} (\parameterVector) =
    \costFunctional_\text{drive} (\parameterVector) +
    \costFunctional_\text{comf} (\parameterVector) +
    \costFunctional_\text{coll} (\parameterVector).
\end{equation}
The first term
\begin{equation} \label{eq:cost_driving_terms}
    \costFunctional_\text{drive} (\parameterVector) =
    \costFunctional_\text{asym} (\VectorSym{\speed}_{i})
\end{equation}
aims to align the speed values $\speed$ in the state vector $\physicalState$ with the desired travel speed $\speed_{\text{des}}$, which is set to 90\% of the speed limit.
Speeds exceeding the maximum allowed are penalized with the quadratic loss, while slower speeds, as common in dense traffic, are penalized with the tolerant loss.
The second term
\begin{align} \label{eq:cost_comfort_terms}
    \costFunctional_\text{comf} (\parameterVector)  =
    \costFunctional_\text{val} & (\controlInput_\acceleration) +
    \costFunctional_\text{val} (\controlInput_\text{lat}) +
    \costFunctional_\text{val} (\boldsymbol{\jerk})
    \nonumber                                                    \\
                               & +
    \costFunctional_\text{ran} (\controlInput_\acceleration) +
    \costFunctional_\text{ran} (\controlInput_\text{lat}) +
    \costFunctional_\text{ran} (\boldsymbol{\jerk})
\end{align}
ensures comfortable driving.
Here, $\controlInput_\acceleration$ represents the longitudinal accelerations, $\controlInput_\text{lat}$ represents the lateral accelerations (calculated using the vehicle model), and $\boldsymbol{\jerk}$ represents the jerk values (calculated from $\controlInput_\acceleration$ using finite differences).
To adjust the vehicle's handling characteristics, $\controlInput_\text{lat}$ is penalized separately from $\controlInput_\acceleration$.
For all these comfort term components the desired values are zero.
The third term
\begin{align}
    \costFunctional_\text{coll} (\parameterVector)
    =
    \weight_\text{coll}
    \sum_{\object \in {\objects_{\mkern-2.5mu\scene}}}
    \Bigg[
     &
    \Bigg(
    {
    {1 + \errorFunction
    {
    \bigg(
    \frac{
    ({\VectorSym{\positionFrenetLon}} - {\VectorSym{\positionFrenetLon}_{\object}})
    }{
    \sqrt{2 \lambda_{\text{lon}} ({\covarianceMatrix}_{\object})}
    }
    \bigg)
    }
    }
    }
    \Bigg)  \nonumber \\
     &
    \Bigg(
    {
    {1 + \errorFunction
    {
    \bigg(
    \frac{
    ({\VectorSym{\positionFrenetLat}} - {\VectorSym{\positionFrenetLat}_{\object}})
    }{
    \sqrt{2 \lambda_{\text{lat}} ({\covarianceMatrix}_{\object})}
    }
    \bigg)
    }
    }
    }
    \Bigg)
    \Bigg]
\end{align}
penalizes collision risks with scene objects.
It projects the position uncertainties of each object onto the reference path and calculates the cumulative distribution at the projected ego vehicle's position using \cref{eq:cartesian2frenet,eq:cumulative_distribution_func}, respectively.
The resulting probabilities in longitudinal and lateral direction are multiplied and then summed over the planning horizon.

\subsection{Model Predictive Contouring Control (MPCC)}
\label{sec:mpcc}

Instead of repeatedly calculating the longitudinal position $\positionFrenetLon$ for the purpose of analyzing interactions and traffic rules, the MPCC formulation \cite{schwarting2017safe, liniger2015optimization} directly integrates it into the state $\physicalState_{\text{mpcc}} = {[\physicalState, \positionFrenetLon]}$, while incorporating its derivative -- the projected speed along the route -- into the action variables $\controlInput_{\text{mpcc}} = {[\controlInput, \dot{\positionFrenetLon}]}$.
The formulation first determines the reference Cartesian coordinates and the corresponding yaw angle $\big( \positionCartesianX^{\text{ref}} (\positionFrenetLon_i), \positionCartesianY^{\text{ref}} (\positionFrenetLon_i), \yawAngle^{\text{ref}} (\positionFrenetLon_i) \big)$ for the desired longitudinal positions.
Subsequently, as part of the optimization process, it computes the longitudinal and lateral contouring errors
\begin{subequations} \label{eq:mpcc_error}
    \begin{align}
        \error_{\text{lon}}
         & =
        - (\positionCartesianX^{\text{ref}} - \positionCartesianX) \sin (\yawAngle^{\text{ref}})
        + (\positionCartesianY^{\text{ref}} - \positionCartesianY) \cos (\yawAngle^{\text{ref}}) \label{eq:mpcc_error_lon} \\
        \error_{\text{lat}}
         & =
        (\positionCartesianX^{\text{ref}} - \positionCartesianX) \cos (\yawAngle^{\text{ref}})
        + (\positionCartesianY^{\text{ref}} - \positionCartesianY) \sin (\yawAngle^{\text{ref}}). \label{eq:mpcc_error_lat}
    \end{align}
\end{subequations}
These errors are calculated within a coordinate frame that aligns with the tangent vector at the reference coordinates, as illustrated in \cref{fig:mpcc}.
The deviations build up the costs
\begin{equation} \label{eq:cost_mpcc_terms}
    \costFunctional_\text{mpcc} (\parameterVector) =
    \costFunctional_\text{val} (\VectorSym{\error}_{\text{lon}}) +
    \costFunctional_\text{val} (\VectorSym{\error}_{\text{lat}}) +
    \costFunctional_\text{val} (\dot{\VectorSym{\positionFrenetLon}}).
\end{equation}
Notably, these cost terms complement the loss term defined in \cref{eq:cost_driving_terms}.

\begin{figure}[th!]
    \centering
    \def\svgwidth{\columnwidth}
    \includesvg{1_figures/mpcc/mpcc}
    \caption{
        Uncertainty projection in conjunction with the MPCC formulation.
        The ego vehicle, depicted in blue, is modeled using three circles.
        Its position is projected onto the closest centerline segment.
        Lateral and longitudinal errors between the reference contouring point and the projected point are denoted with $\error_{\text{lat}}$ and $\error_{\text{lon}}$, respectively.
        The uncertainty of the other vehicle's position is modeled by an ellipse and is projected onto the closest centerline segment.
    }
    \label{fig:mpcc}
\end{figure}

With the MPCC formulation in place, we augment the total cost function outlined in \cref{eq:cost_total} as
\begin{equation} \label{eq:cost_decision_theoretic}
    \costFunctional_\text{total} (\parameterVector)
    =
    \costFunctional_\text{drive} (\parameterVector)
    +
    \costFunctional_\text{comf} (\parameterVector)
    +
    \costFunctional_\text{coll} (\parameterVector)
    +
    \costFunctional_\text{mpcc} (\parameterVector).
\end{equation}
In addition, the formulation imposes further constraints on $\VectorSym{\error}_{\text{lon}}$ and $\VectorSym{\error}_{\text{lat}}$, confining them within an interval defined by upper and lower bounds.
It is worth emphasizing that, while the reference coordinates and tangent vectors in the MPCC formulation serve for error calculation, our approach further leverages them for projecting uncertainties.

\subsection{Incorporating Maneuver Preferences in MPC Framework}

Planning with unclear maneuver intentions from others introduces a high level of uncertainty, as previously discussed.
Particularly for continuous optimization based planners, the challenge lies in the necessity of committing to a single maneuver, or homotopy, before optimizing motion within it.
A practical alternative is to consider all maneuvers simultaneously with their preference scores, or weights, and plan a motion accordingly.
This approach enables the planner to optimize a motion profile that is not confined to a single homotopy, but rather, is \emph{homotopy-free} or \mbox{\emph{maneuver-neutral} \cite{tas2018decision,tas2020tackling}}. 

We augment the parameter vector of the planning problem by incorporating state and action parameters that correspond to distinct available maneuvers.
In the case of two maneuvers A and B, this modification yields the parameter vector
\begin{equation} \label{eq:parameters_homotopy_free_planning}
    \parameterVector^\star =
    [
    \underbrace{ \parameterVector_0, \, \parameterVector_{1:\pinnedIndices} }_{\text{pinned}}, \,
    \underbrace{\overbrace{\parameterVector_{\pinnedIndices + 1 : 2\pinnedIndices}}^\mathrm{C}}_{ \text{executed next}}, \,
    \underbrace{
        \overbrace{\parameterVector_{2\pinnedIndices + 1: n}}^\mathrm{CA}, \,
        \overbrace{\parameterVector_{n + 1 : 2n - 2\pinnedIndices}}^{\mathrm{CB}}
    }_\text{will be recalculated} \,
    ]. %
\end{equation}
The braces C, CA, and CB represent the segments of the maneuver on the path-time diagram shown in \cref{fig:phantom_object_treatment}.
It should be noted that this planning scheme requires \mbox{$(n - 2 \pinnedIndices) ( \text{dim}(\physicalState)  + \text{dim}(\controlInput))$} additional parameters for each extra maneuver considered.
However, in many situations, the number of alternative maneuver options can be reduced to two.


Despite modifications to the parameter vector, we preserve the original form of both the cost and constraint functions.
We, therefore, feed parameters into these functions such that they represent individual maneuvers.
Given two distinct maneuver options A and B, the schema for parameter passing is
\begin{subequations}
    \begin{align}
        \parameterVector^\star_\text{A} & =
        {
        \left[
            \parameterVector_{0:\pinnedIndices},
            \parameterVector_{\pinnedIndices + 1:2\pinnedIndices},
            \parameterVector_{2\pinnedIndices + 1:n}
            \right]
        },                                  \\
        \parameterVector^\star_\text{B} & =
        {
        \left[
            \parameterVector_{0:\pinnedIndices},
            \parameterVector_{\pinnedIndices + 1:2\pinnedIndices},
            \parameterVector_{n + 1:2n - 2\pinnedIndices}
            \right]
        }.
    \end{align}
\end{subequations}
We compute the total cost as a weighted sum
\begin{align} \label{eq:parameters_homotopy_free_planning_cost}
    \costFunctional^\star_\text{total} =
    \weight_\text{A} \, \costFunctional_\text{total}
    \left(
    \parameterVector^\star_\text{A}
    \right)
    +
    \weight_\text{B} \, \costFunctional_\text{total}
    \left(
    \parameterVector^\star_\text{B}
    \right).
\end{align}
The weights $\weight_{(\cdot)}$ represent the \textit{maneuver preferences} and are determined by a decision function $F_{\text{DT}}$, which evaluates the current scene.
$F_{\text{DT}}$ can represent a model-based or a learning-based approach.
In this work, we do not delve into the design of the optimal decision function, but use the maneuver probabilities or the existence probabilities of other participants as maneuver weights.

In some scenarios, choosing between maneuver options ${\mathrm{A}}$ and ${\mathrm{B}}$ may involve trade-offs.
Let $\VectorSym{\positionFrenetLon}_{\mathrm{A}}$ and $\VectorSym{\positionFrenetLon}_{\mathrm{B}}$ represent the longitudinal motion profiles for options ${\mathrm{A}}$ and ${\mathrm{B}}$, respectively.
Suppose that motion profiles $\VectorSym{\positionFrenetLon}_{\mathrm{A}}$ and $\VectorSym{\positionFrenetLon}_{\mathrm{B}}$ obtained by a conventional planner are safe $\forall i \in \{k+ 1, \ldots, 2k\}$.
Then, there exists a motion profile ${\mathrm{C}}$, such that $\VectorSym{\positionFrenetLon}_{\mathrm{B}, i} \leq \VectorSym{\positionFrenetLon}_{\mathrm{C}, i} \leq \VectorSym{\positionFrenetLon}_{\mathrm{A}, i}$, that is also safe $\forall i \in \{\pinnedIndices + 1, \ldots, 2\pinnedIndices\}$, under appropriate constraints.
We will detail these constraints in the following subsection.

It is important to underscore that the objective function presented in \cref{eq:parameters_homotopy_free_planning_cost} enables the planner to postpone decisions, taking advantage of the expectation that more information will be available in the future. %
Consequently, we term this capability as \textit{passive} information gathering.

\subsection{Decision-Theoretic MPC}

The outlined planning approach with maneuver preferences in the preceding subsection forms the core of \emph{decision-theoretic MPC}.
We introduce additional constraints on this core formulation to maintain safety and to better guide the solver against conflicting objectives and uncertain or incomplete information.
In our application, this results in
\begin{subequations}
    \begin{align}
        \min_{\parameterVector^\star}
        \quad
         &
        \weight_\text{A} \, \costFunctional_\text{total}
        \left(
        \parameterVector^\star_\text{A}
        \right)
        +
        \weight_\text{B} \, \costFunctional_\text{total}
        \left(
        \parameterVector^\star_\text{B}
        \right)
        \\
        \text{s.t.\ } \quad
         & \equalityConstraints_{\text{dyn}} (\parameterVector^\star_{\text{A}}) = \VectorSym{0},                                                   \\
         & \equalityConstraints_{\text{dyn}} (\parameterVector^\star_{\text{B}}) = \VectorSym{0},                                                   \\
         & \inequalityConstraints_{\text{acc}} (\parameterVector^\star_{\text{A}}) \leq \VectorSym{0},                                              \\
         & \inequalityConstraints_{\text{acc}} (\parameterVector^\star_{\text{B}}) \leq \VectorSym{0},                                              \\
         & \inequalityConstraints_{\text{cnt}} (\parameterVector^\star_{\text{A}}) \leq \VectorSym{0},                                              \\
         & \inequalityConstraints_{\text{cnt}} (\parameterVector^\star_{\text{B}}) \leq \VectorSym{0},                                              \\
         & \inequalityConstraint_{\text{crc}} (\parameterVector^\star_i) \leq 0, \quad \;\;\,\, \forall i \in \{0, \ldots, 2\pinnedIndices \}       \\
         & \inequalityConstraint_{\text{chn}} (\parameterVector^\star_i) \leq 0. \quad \hspace{3.2mm} \forall i \in \{0, \ldots, 2\pinnedIndices \}
    \end{align}
    \label{equ:decision_theoretic_opt_probl}
\end{subequations}
Here, the objective function directly employs \cref{eq:parameters_homotopy_free_planning_cost}, which is a weighted sum of the costs in \cref{eq:cost_decision_theoretic}.
The constraint
$\equalityConstraints_{\text{dyn}}$ represents the vehicle motion models referenced in \cref{sec:vehicle_model},
$\inequalityConstraints_{\text{acc}}$ denotes longitudinal and lateral acceleration limits,
$\inequalityConstraints_{\text{cnt}}$ denotes the longitudinal and lateral contouring error constraints introduced in \cref{sec:mpcc},
$\inequalityConstraint_{\text{crc}}$ denotes the circle-to-ellipse safety constraint defined in \cref{eq:collision_ellipse}.
Lastly, $\inequalityConstraint_{\text{chn}}$ represents the chance constraints, composed of
\begin{equation}
    \inequalityConstraint_\text{chn} (\parameterVector)
    =
    \inequalityConstraint_\text{stop} (\parameterVector)
    \wedge
    \inequalityConstraint_\text{vis} (\parameterVector)
    \wedge
    \inequalityConstraint_\text{yield} (\parameterVector) \, ,
\end{equation}
where its components are defined in \cref{eq:chance_constraints_braking_delta,eq:chance_constraints_braking,eq:chance_constraints_braking_MP_general}.
We name this planning formulation a {decision-theoretic MPC}\footnote{We refrain from abbreviating our method as DTMPC to avoid confusion with dynamic tube MPC \cite{lopez2019dynamic}.}.

It is important to note that this formulation does not enforce any constraints for obstacle avoidance beyond the index $2\pinnedIndices$.
This is a reasonable approach, given that newly detected objects beyond this index could invalidate the preceding solution used for initialization at the current planning instance, potentially causing the optimization routine to fail.

\section{Solving Local-Optimization Problems}
\label{sec:optimization}

The optimization problem presented is not only complex, but it also must be solved in real-time.
For this reason, we use an Interior Point Method (IPM), a type of second-order gradient-based nonlinear optimization algorithm \cite{kochenderfer2019algorithms, wright2005interior, nocedal2006numerical}.
IPMs transform inequality constraints into equality constraints by introducing slack variables and define a logarithmic barrier term to satisfy these constraints.
As the optimization process progresses, the value of the barrier term is reduced until the Karush-Kuhn-Tucker conditions for an equality-constrained problem are met.
This procedure makes IPMs robust and eliminates the need for feasible initialization.

We use the IPM solver \textsc{Ipopt} \cite{wachter2006implementation}, known as the most powerful and robust nonlinear solver available.
It is a primal-dual algorithm and features a ``filter'' line search method, a second-order correction method for the step factor selection, and inertia correction capabilities which enable its robust operation.
Whenever it cannot find a feasible line search step factor, it enters into restoration mode and minimizes the constraint violation.

The presented optimization problem has many optimization parameters and constraints, thereby making the calculation of gradients and Hessians more challenging.
Various methods for computing gradients and Hessians are compared in \cref{tab:derivatives_comparison}.
We choose automatic differentiation for its flexibility and efficiency \cite{baydin2018automatic}.
Among the automatic differentiation libraries available for the C++ programming language, CppAD \cite{cppad} and CasADi \cite{casadi} are the most powerful.
They both employ sparse-matrix methods to reduce the storage and computational overhead.
The main of advantage of CppAD over CasADi lies in its ability to be integrated into templated function definitions, thus blending into existing software without requiring a complete rewrite.
We assess the performance of both libraries in our benchmarks.

\begin{table}[H]
\centering

\begingroup
\sisetup{
    table-number-alignment = center,
    table-figures-integer = 1,
    table-figures-decimal = 2}

\begin{tabular}{l*{6}{c}r}
    \toprule
    {Derivative Method}   & {Accuracy} & {Speed} & {Setup Time} & {Error Safe} \\
    \cmidrule(lr){1-1} \cmidrule(lr){2-5}
    Analytical Deriv.\    & +++        & ++      & -            & -            \\
    Numeric Diff.\        & -          & -       & +++          & +++          \\
    Symbolic Diff.\       & +++        & +       & +            & ++           \\
    Automatic Diff.\      & +++        & +       & ++           & ++           \\
    Hybrid Auto.\ Diff.\  & +++        & +++     & ++           & ++           \\
    \bottomrule
\end{tabular}
\endgroup

\caption{
    Comparison of various differentiation methods \cite{adrlCT, neunert2016fast}.
}
\label{tab:derivatives_comparison}
\end{table}

From the comparison presented in the table above, it is apparent that hybrid automatic differentiation, \eg CppADCodeGen \cite{cppadcg}, shows better performance over standard automatic differentiation.
This is because CppAD operates on scalars rather than matrices and does not benefit from advanced compiler optimization techniques.
Although this results in a slightly worse performance, the generated code is dependency-free, facilitating its operation on embedded systems and GPUs.
Additionally, it allows static memory allocation, enabling hard real-time guarantees \cite{adrlCT, neunert2016fast}.

\section{Evaluation}
\label{sec:evaluation}

We demonstrate the efficacy of our motion planner in a multitude of driving scenarios, including both simulation studies and driving experiments.
In our evaluations, we set $\samplingInterval = \SI{100}{\milli\second} $ and $\horizon = \SI{6}{\second}$.
Additionally, we set the number of pinned control inputs $\pinnedIndices$ to $2$, except in the phantom object scenario, where we use a value of $4$.

\subsection{Limited Field of View}

To evaluate safe intersection crossing under impaired perception, we examine a scenario where a vehicle's visible field-of-view is obstructed by surrounding buildings (see \cref{fig:limited_visibility_eval}).
Initially, the vehicle is unaware of the intersection at which it must yield to vehicles approaching from other directions.
We analyze speed profiles for various sensor ranges on a path-speed diagram and define the center of the intersection as the origin of the path axis.

\begin{figure}[!ht]
    \vspace{1mm}
    \begin{subfigure}[b!]{0.45\columnwidth}
        \hspace{2mm}
        \def\svgwidth{0.9\columnwidth}
        \fontsize{7}{8}\selectfont
        \includesvg[width=1.04\columnwidth]{1_figures/limited_visibility/map.svg}
    \end{subfigure} \hspace{4mm}
    \begin{subfigure}[b!]{0.45\columnwidth}
        \centering
        \def\svgwidth{0.9\columnwidth}
        \fontsize{7}{8}\selectfont
        \includesvg[width=\columnwidth]{1_figures/limited_visibility/position_velocity.svg}
    \end{subfigure}
    \caption{Trajectories at an intersection crossing for various sensor ranges.}
    \label{fig:limited_visibility_eval}
\end{figure}

Speed profiles indicate that with longer sensor ranges, the vehicle can detect the intersection sooner, enabling a more comfortable reduction in speed.
On the other hand, with shorter sensor ranges, the vehicle identifies the intersecting route closer to the intersection and later in time, resulting in a harsher deceleration.
Note that all profiles on the diagram are of the same length, as the simulation is run for the same time period.
Once the vehicle has enough visible field in the intersection area, all profiles accelerate and reach a peak speed at approximately \SI{20}{\meter} past the center of the intersection.
Afterwards, they decelerate due to the limited field-of-view resulting from the presence of buildings and the curvature of the road.
This implies that sensor ranges exceeding \SI{50}{\meter} may not always be advantageous in urban driving.

\subsection{Postponing Decisions in Obstacle Avoidance}

We evaluate the motion of our decision-theoretic MPC in certain challenging scenarios.
These include an S-shaped swerve scenario and a \mbox{180\textdegree-cornering}, each with objects divided into two groups (see~\cref{fig:homotopy_free_s} and \cref{fig:homotopy_free_cornering} in the appendix).
Only one of the groups actually exists in reality, but this is unknown to the planner.
Therefore, the vehicle has no preference on the corresponding maneuver options and weights them equally.
The resulting motion is optimized for both options in the immediate time horizon, and it maintains a sufficient distance from all obstacles.
Having identical motion in the immediate time horizon across different maneuvers is a distinguishing feature of the decision-theoretic MPC.

\subsection{Chance-Constrained Contingency Planning}

In some situations, environmental constraints might prevent the execution of swerve maneuvers, leaving braking as the only viable option.
Such a situation is illustrated in \cref{fig:phantom_object_treatment}, where the vehicle detects a vehicle with a \SI{50}{\percent} existence probability, moving at \SI{2.0}{\metre\per\second}, located \SI{15.0}{\metre} ahead.
We evaluate two variations of this scenario: in the first, the phantom detection is resolved after $\SI{0.3}{\second}$, while in the second, its existence remains unclear.
The comparison between the decision-theoretic MPC and a conventional planner is depicted in \cref{fig:eval_decision_theoretic_planning}.
While the conventional planner calculates the motion by assuming the phantom to be valid, the decision-theoretic planner assesses the likelihood of a phantom presence and assigns $\weight_\text{A} = 0.5$ and $\weight_\text{B} = 0.5$ to the braking and driving straight maneuver options, respectively.
By ensuring the feasibility of a fallback maneuver, it reacts more comfortably to the phantom without compromising safety.
The slight jump in the left subfigure results from the change of the reference contour in the MPCC formulation.

\begin{figure}[h!]
    \centering
    \vspace{2mm}
    \begin{subfigure}[t]{0.49\columnwidth}
        \centering
        \fontsize{7}{8}\selectfont
        \begin{tikzpicture}[>=stealth]
    
    \begin{axis}[
            width=4.8cm,
            height=3.5cm,
            set layers,
            axis line style={on layer=axis foreground},
            axis lines=middle,
            xmin=0.0, xmax=3.0,     
            ymin=8.0, ymax=10.4, 
            tick align=outside,
            minor x tick num=4,
            xtick={0.0, 0.5, 1.0, 1.5, 2.0, 2.5},
            extra x ticks={0},
            extra x tick labels={0},
            minor y tick num=1,
            xlabel={$t$ [s]},
            ylabel={$v$ [m/s]},
            style= thick     
        ]
        

        \addplot [semithick, domain=0:20, mark=none, myBlue] file {1_figures/phantom_eval/data/speed_400ms_conventional.dat};
        \addplot [semithick, domain=0:20, mark=none, myRed] file {1_figures/phantom_eval/data/speed_400ms_neutral.dat};

    \end{axis}
\end{tikzpicture}
        \caption{
            \mbox{Phantom cleared at $t=0.3$}.
        }
        \label{fig:eval_decision_theoretic_planning_cleared}
    \end{subfigure} \hfill
    \begin{subfigure}[t]{0.49\columnwidth}
        \centering
        \fontsize{7}{8}\selectfont
        \begin{tikzpicture}[>=stealth]
    
    \begin{axis}[
            width=4.8cm,
            height=3.5cm,
            set layers,
            axis line style={on layer=axis foreground},
            axis lines=middle,
            xmin=0.0, xmax=9.8,     
            ymin=0.0, ymax=12.0, 
            tick align=outside,
            minor x tick num=4,
            xtick={0.0, 1.0, 2.0, 3.0, 4.0, 5.0, 6.0, 7.0, 8.0},
            extra x ticks={0},
            extra x tick labels={0},
            minor y tick num=1,
            xlabel={$t$ [s]},
            ylabel={$v$ [m/s]},
            style= thick     
        ]
        
        
        \addplot [semithick, domain=0:20, mark=none, myBlue] file {1_figures/phantom_eval/data/speed_extra_long_conventional.dat};
        \addplot [semithick, domain=0:20, mark=none, myRed] file {1_figures/phantom_eval/data/speed_extra_long_neutral.dat};

    \end{axis}
\end{tikzpicture}
        \caption{
            Phantom detection persists.
        }
        \label{fig:eval_decision_theoretic_planning_persistent}
    \end{subfigure}
    \caption{
        Executed speed profiles between the decision-theoretic MPC \mbox{(\hspace{-1.4mm}\continuousline{myRed}\hspace{-1.4mm})} and a conventional MPC \mbox{(\hspace{-1.4mm}\continuousline{myBlue}\hspace{-1.4mm})} in the phantom object detection scenario.
        While the conventional planner calculates the motion by assuming the phantom to be valid, the decision-theoretic planner assesses the likelihood of a phantom presence.
    }
    \label{fig:eval_decision_theoretic_planning}
\end{figure}

\subsection{Driving Experiments with Rule-violating Participants}

\begin{figure*}

    \begin{subfigure}[b!]{0.24\textwidth}
        \centering
        \frame{\includegraphics[scale=0.12]{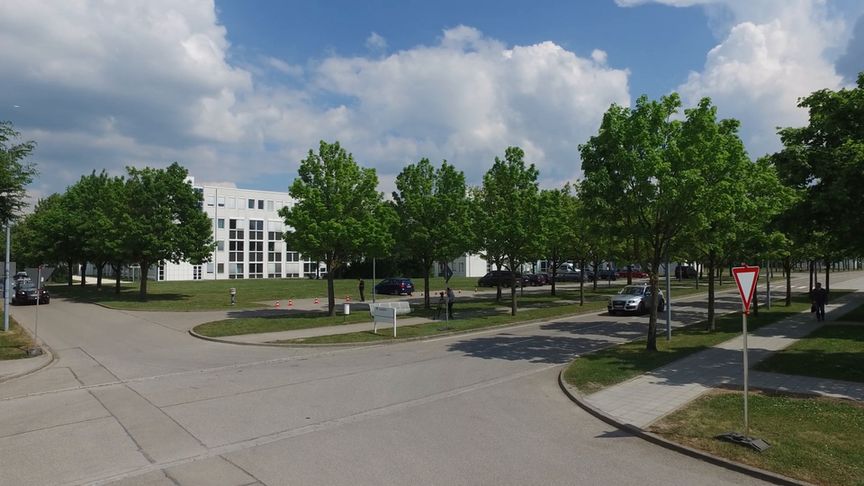}}
        \vspace{-1mm}
        \caption*{Frame 1}
        \vspace{1mm}
    \end{subfigure} \hfill
    \begin{subfigure}[b!]{0.24\textwidth}
        \centering
        \frame{\includegraphics[scale=0.12]{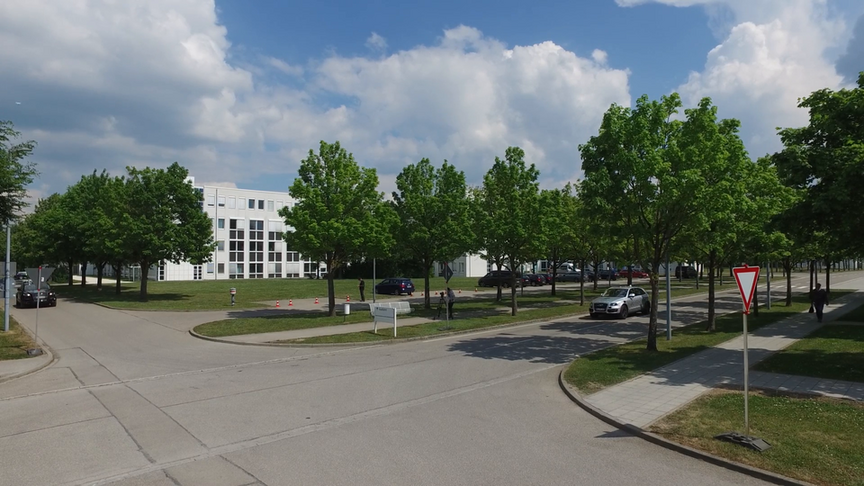}}
        \vspace{-1mm}
        \caption*{Frame 2}
        \vspace{1mm}
    \end{subfigure} \hfill
    \begin{subfigure}[b!]{0.24\textwidth}
        \centering
        \frame{\includegraphics[scale=0.12]{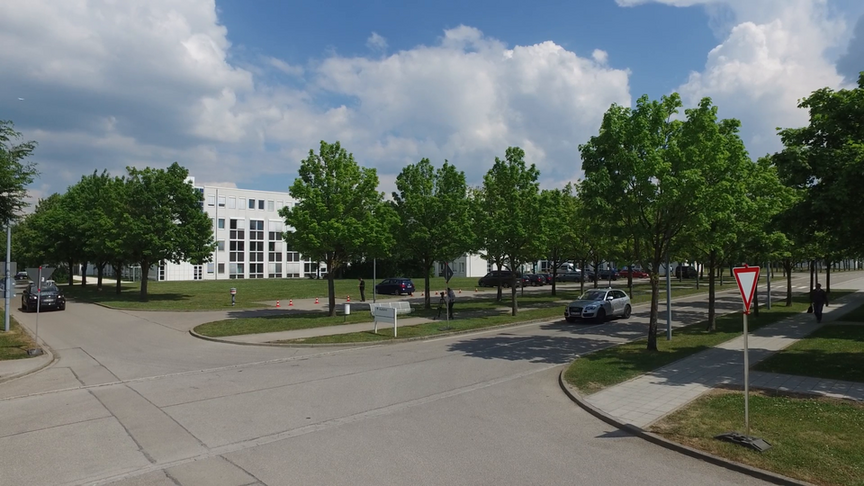}}
        \vspace{-1mm}
        \caption*{Frame 3}
        \vspace{1mm}
    \end{subfigure}  \hfill
    \begin{subfigure}[b!]{0.24\textwidth}
        \centering
        \frame{\includegraphics[scale=0.12]{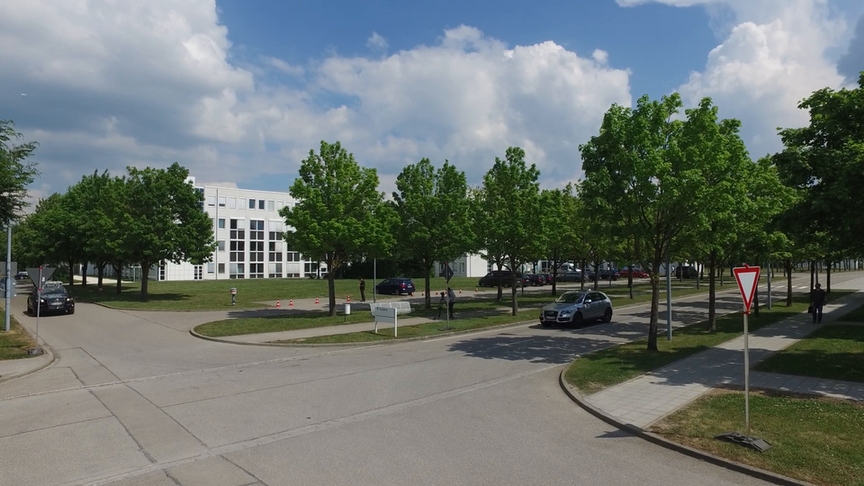}}
        \vspace{-1mm}
        \caption*{Frame 4}
        \vspace{1mm}
    \end{subfigure}

    \begin{subfigure}[b!]{0.24\textwidth}
        \centering
        \frame{\includegraphics[scale=0.12]{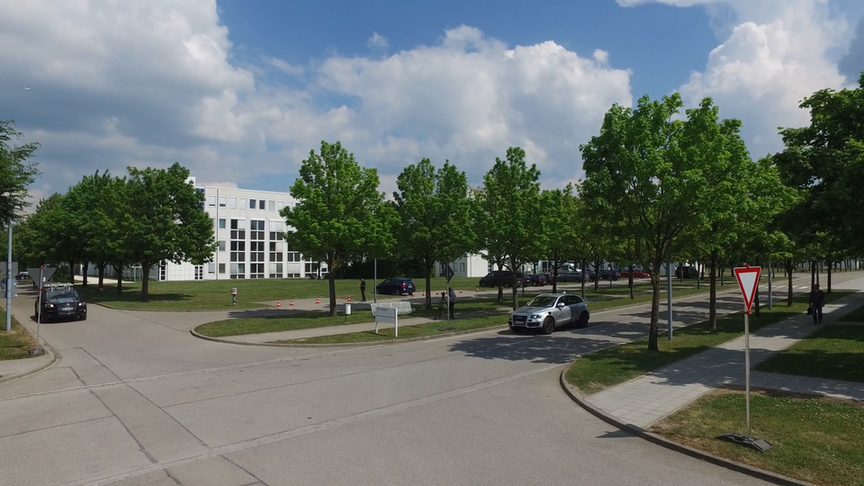}}
        \vspace{-1mm}
        \caption*{Frame 5}
        \vspace{1mm}
    \end{subfigure} \hfill
    \begin{subfigure}[b!]{0.24\textwidth}
        \centering
        \frame{\includegraphics[scale=0.12]{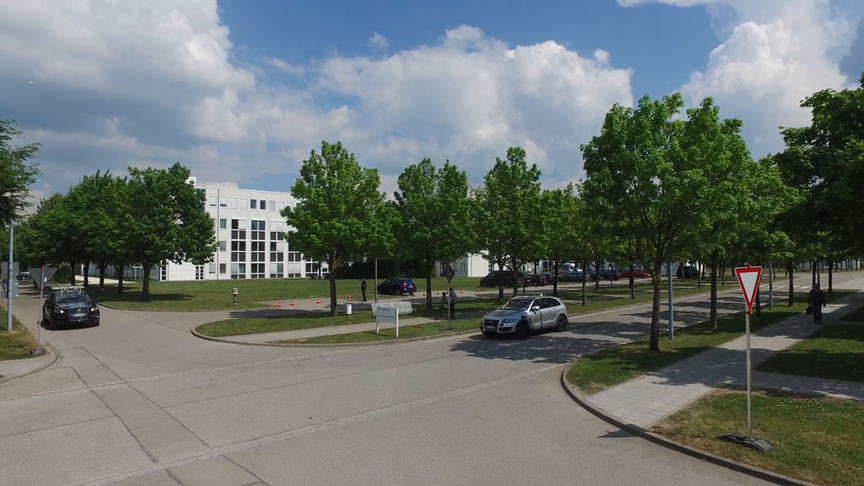}}
        \vspace{-1mm}
        \caption*{Frame 6}
        \vspace{1mm}
    \end{subfigure} \hfill
    \begin{subfigure}[b!]{0.24\textwidth}
        \centering
        \frame{\includegraphics[scale=0.12]{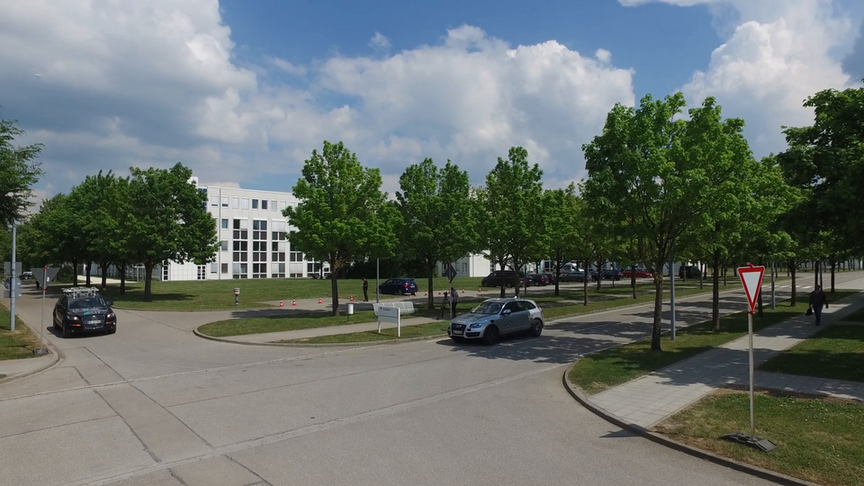}}
        \vspace{-1mm}
        \caption*{Frame 7}
        \vspace{1mm}
    \end{subfigure} \hfill
    \begin{subfigure}[b!]{0.24\textwidth}
        \centering
        \frame{\includegraphics[scale=0.12]{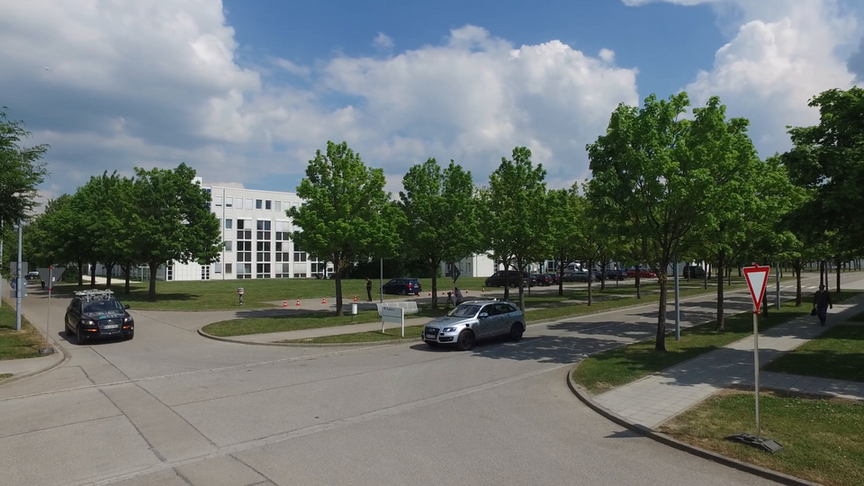}}
        \vspace{-1mm}
        \caption*{Frame 8}
        \vspace{1mm}
    \end{subfigure}

    \begin{subfigure}[b!]{0.24\textwidth}
        \centering
        \frame{\includegraphics[scale=0.12]{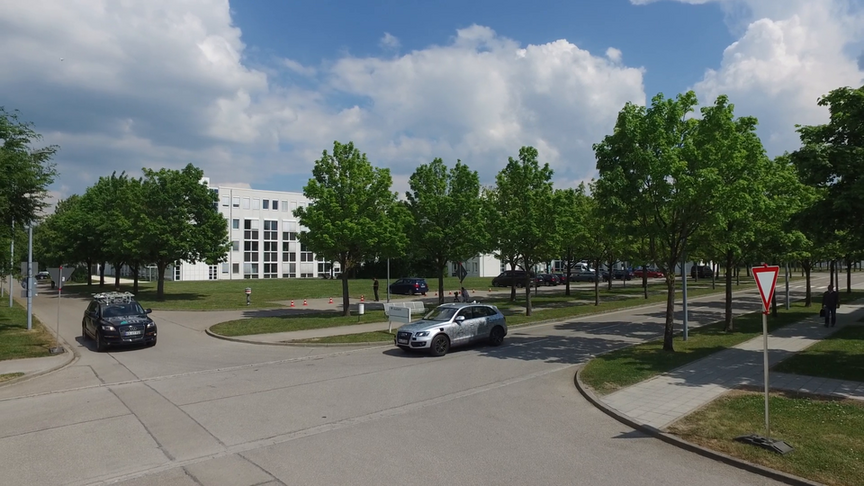}}
        \vspace{-1mm}
        \caption*{Frame 9}
        \vspace{1mm}
    \end{subfigure} \hfill
    \begin{subfigure}[b!]{0.24\textwidth}
        \centering
        \frame{\includegraphics[scale=0.12]{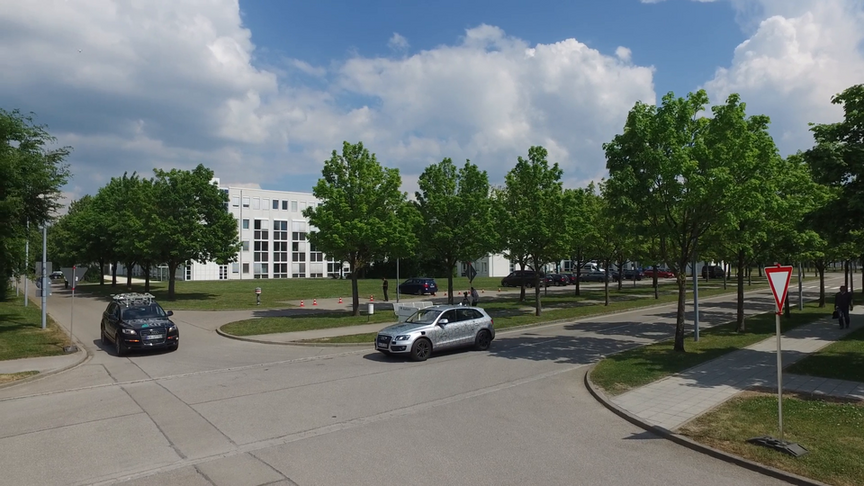}}
        \vspace{-1mm}
        \caption*{Frame 10}
        \vspace{1mm}
    \end{subfigure} \hfill
    \begin{subfigure}[b!]{0.24\textwidth}
        \centering
        \frame{\includegraphics[scale=0.12]{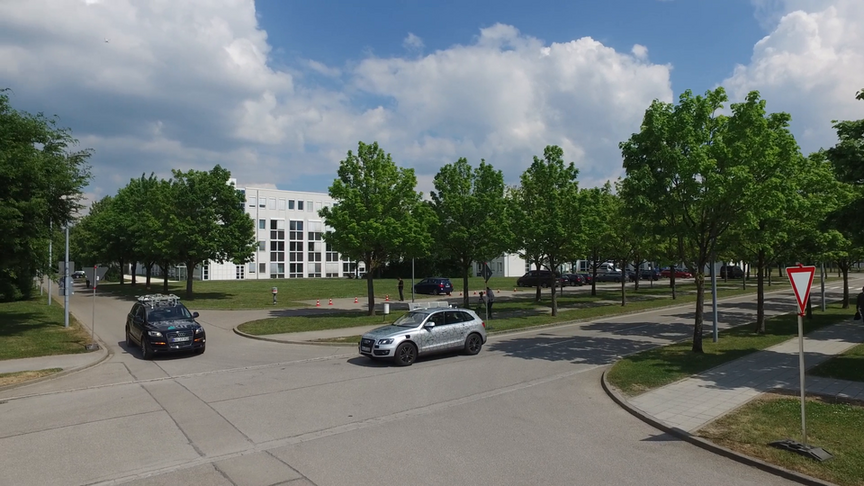}}
        \vspace{-1mm}
        \caption*{Frame 11}
        \vspace{1mm}
    \end{subfigure} \hfill
    \begin{subfigure}[b!]{0.24\textwidth}
        \centering
        \frame{\includegraphics[scale=0.12]{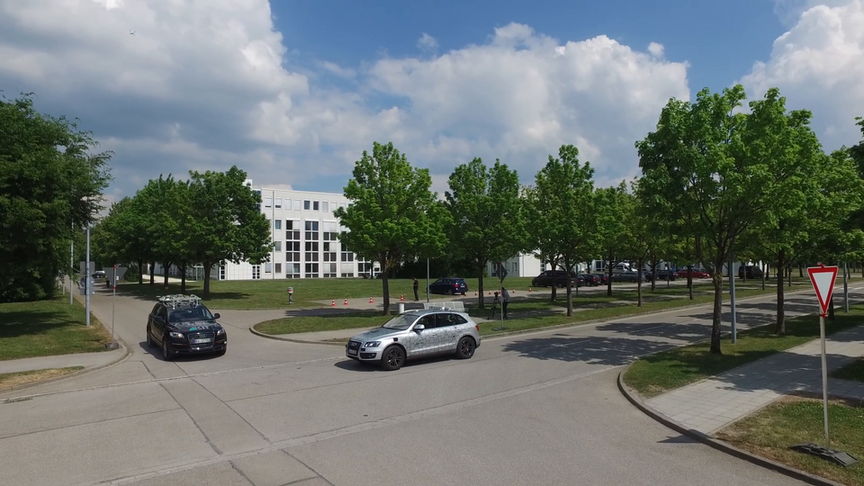}}
        \vspace{-1mm}
        \caption*{Frame 12}
        \vspace{1mm}
    \end{subfigure}

    \begin{subfigure}[b!]{0.24\textwidth}
        \centering
        \frame{\includegraphics[scale=0.12]{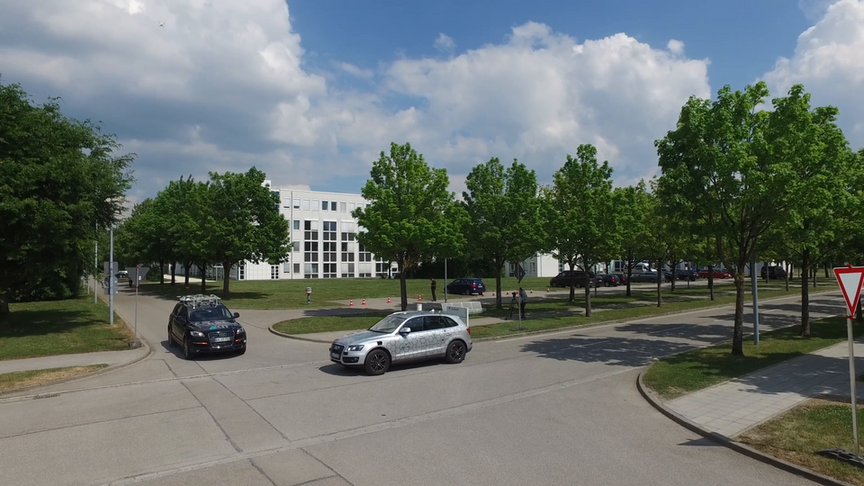}}
        \vspace{-1mm}
        \caption*{Frame 13}
        \vspace{1mm}
    \end{subfigure} \hfill
    \begin{subfigure}[b!]{0.24\textwidth}
        \centering
        \frame{\includegraphics[scale=0.12]{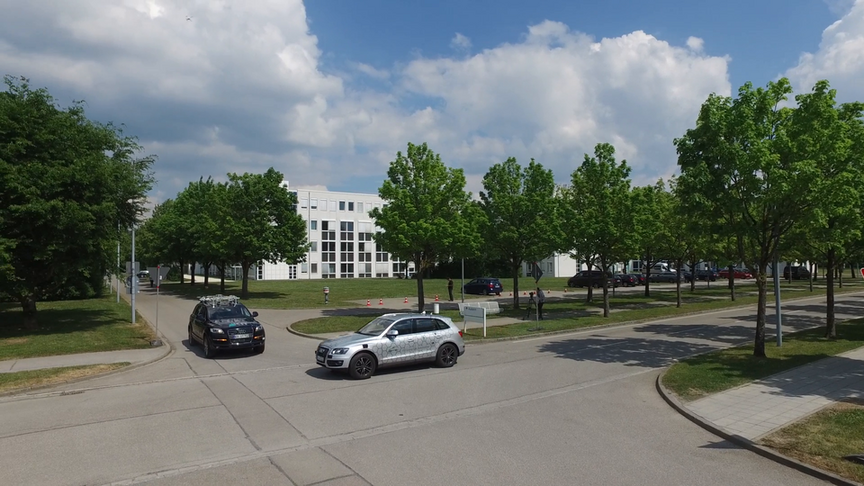}}
        \vspace{-1mm}
        \caption*{Frame 14}
        \vspace{1mm}
    \end{subfigure} \hfill
    \begin{subfigure}[b!]{0.24\textwidth}
        \centering
        \frame{\includegraphics[scale=0.12]{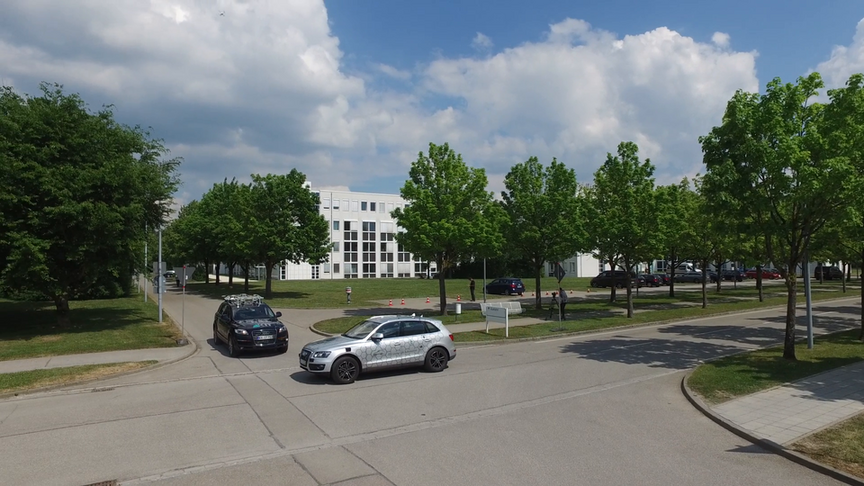}}
        \vspace{-1mm}
        \caption*{Frame 15}
        \vspace{1mm}
    \end{subfigure} \hfill
    \begin{subfigure}[b!]{0.24\textwidth}
        \centering
        \frame{\includegraphics[scale=0.12]{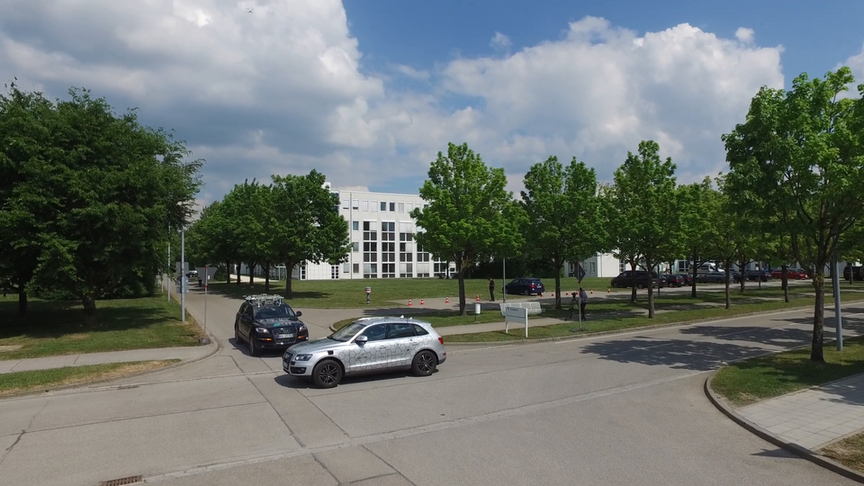}}
        \vspace{-1mm}
        \caption*{Frame 16}
        \vspace{1mm}
    \end{subfigure}
    \caption{
        Safe intersection crossing in the presence of rule-violating participants.
        The autonomous gray vehicle has the right of way.
        However, the speed of the oncoming black vehicle suggests it will not yield.
        The autonomous vehicle reduces its speed to satisfy the yield maneuver before the intersection.
        Once it becomes clear that the oncoming vehicle will yield, it accelerates back to its desired driving speed.
        Frames are incremented every $\SI{0.25}{\second}$.
    }
    \label{fig:evaluation_uncompliant_demo_camera}
\end{figure*}

Our decision-theoretic MPC is capable of proactively handling rule-violating behaviors by other vehicles.
This ability is showcased through driving experiments performed at a T-intersection, where the autonomous vehicle has the right-of-way, and an oncoming vehicle must yield to it (see \cref{fig:evaluation_uncompliant_demo_camera}, where the autonomous vehicle, CoCar \cite{kohlhaas2013towards}, is gray, and the oncoming vehicle is black).
To introduce variability in the experiment and prove the robustness of our algorithm, the oncoming vehicle is driven manually.
It approaches the intersection at a speed that suggests it will not yield to the autonomous vehicle, yet brakes at the very last moment.
To isolate the effect of perception artifacts, the oncoming vehicle transmits its position and speed via V2V-communication to the autonomous vehicle.

\begin{figure*}
    \vspace{-12mm} 
    \begin{subfigure}[b!]{0.24\textwidth}
        \centering
        \frame{\includegraphics[scale=0.137]{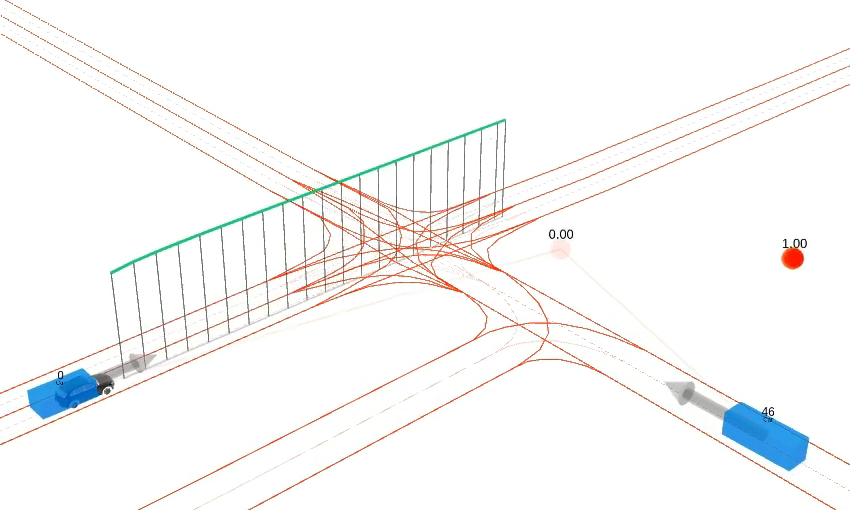}}
        \vspace{-1mm}
        \caption*{Frame 1}
        \vspace{1mm}
        \label{fig:evaluation_uncompliant_demo_ros_a}
    \end{subfigure} \hfill
    \begin{subfigure}[b!]{0.24\textwidth}
        \centering
        \frame{\includegraphics[scale=0.137]{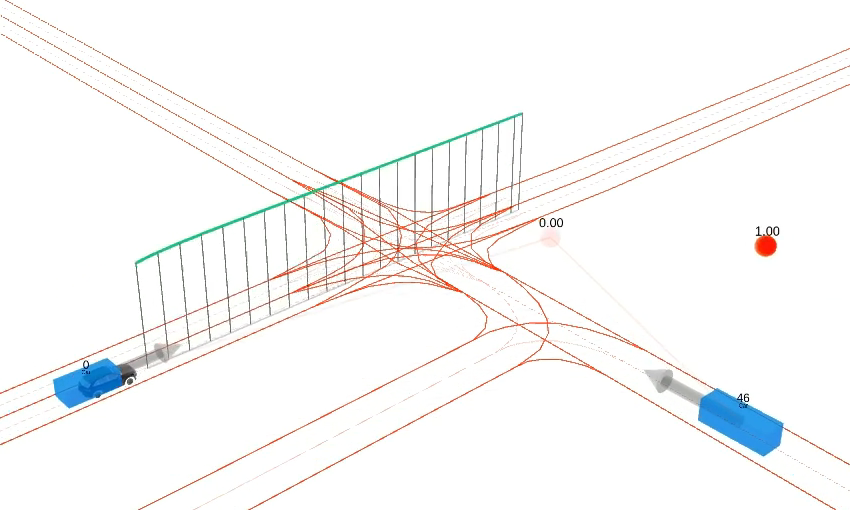}}
        \vspace{-1mm}
        \caption*{Frame 2}
        \vspace{1mm}
        \label{fig:evaluation_uncompliant_demo_ros_b}
    \end{subfigure} \hfill
    \begin{subfigure}[b!]{0.24\textwidth}
        \centering
        \frame{\includegraphics[scale=0.137]{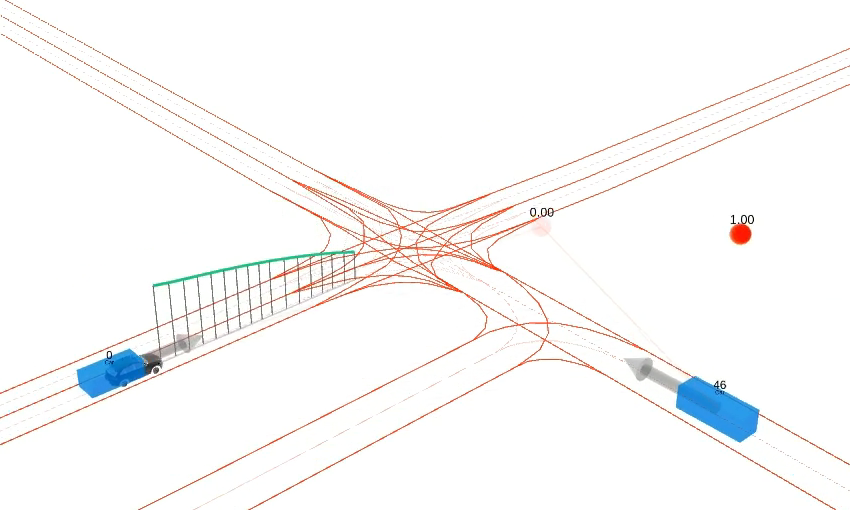}}
        \vspace{-1mm}
        \caption*{Frame 3}
        \vspace{1mm}
        \label{fig:evaluation_uncompliant_demo_ros_c}
    \end{subfigure}  \hfill
    \begin{subfigure}[b!]{0.24\textwidth}
        \centering
        \frame{\includegraphics[scale=0.137]{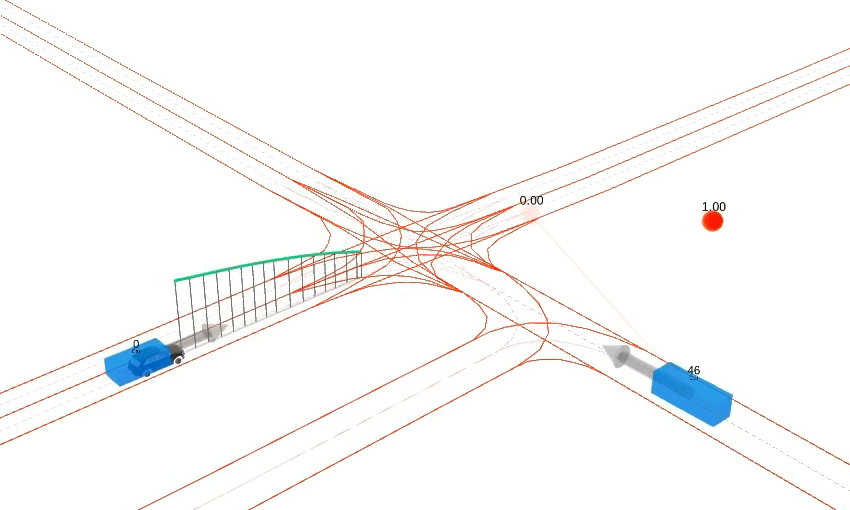}}
        \vspace{-1mm}
        \caption*{Frame 4}
        \vspace{1mm}
        \label{fig:evaluation_uncompliant_demo_ros_d}
    \end{subfigure}

    \begin{subfigure}[b!]{0.24\textwidth}
        \centering
        \frame{\includegraphics[scale=0.137]{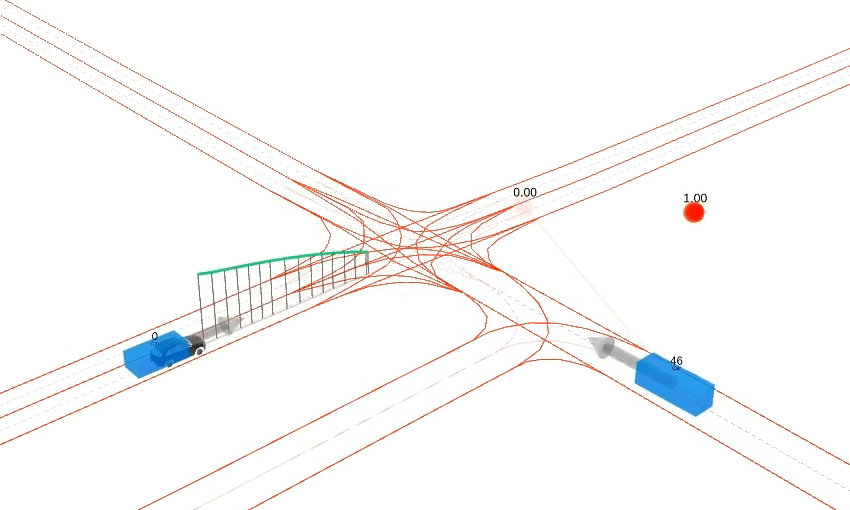}}
        \vspace{-1mm}
        \caption*{Frame 5}
        \vspace{1mm}
        \label{fig:evaluation_uncompliant_demo_ros_e}
    \end{subfigure} \hfill
    \begin{subfigure}[b!]{0.24\textwidth}
        \centering
        \frame{\includegraphics[scale=0.137]{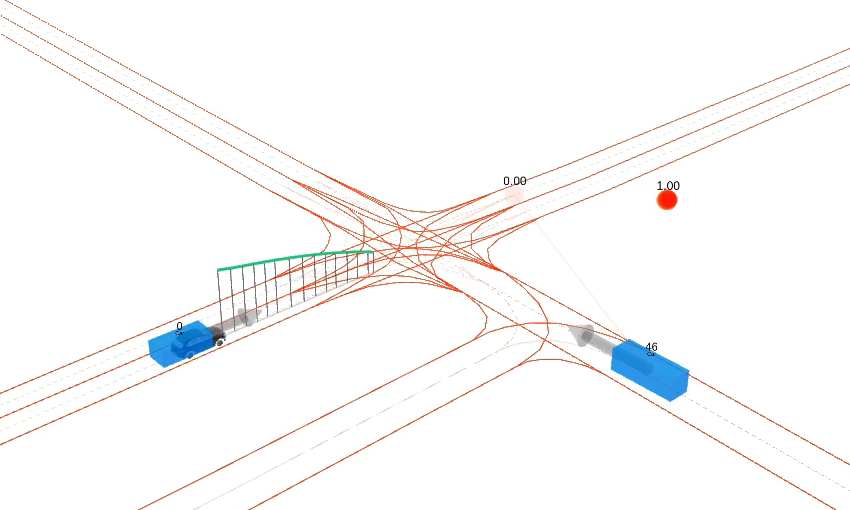}}
        \vspace{-1mm}
        \caption*{Frame 6}
        \vspace{1mm}
        \label{fig:evaluation_uncompliant_demo_ros_f}
    \end{subfigure} \hfill
    \begin{subfigure}[b!]{0.24\textwidth}
        \centering
        \frame{\includegraphics[scale=0.137]{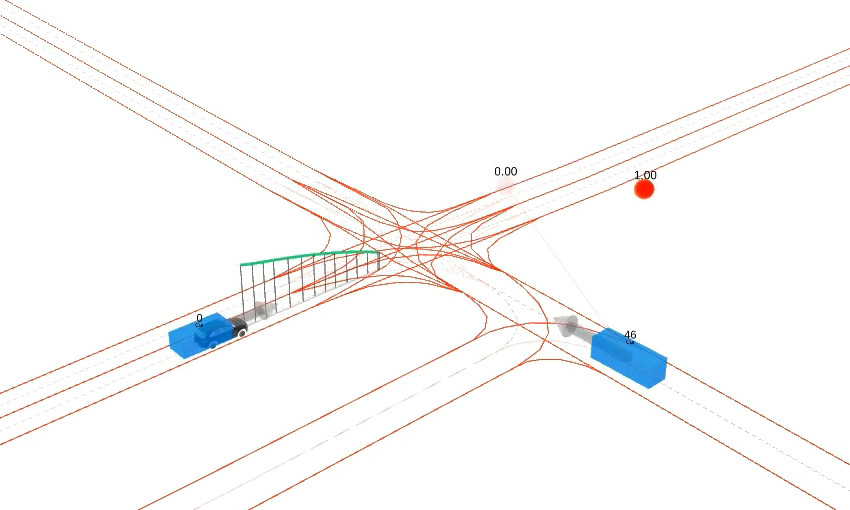}}
        \vspace{-1mm}
        \caption*{Frame 7}
        \vspace{1mm}
        \label{fig:evaluation_uncompliant_demo_ros_g}
    \end{subfigure} \hfill
    \begin{subfigure}[b!]{0.24\textwidth}
        \centering
        \frame{\includegraphics[scale=0.137]{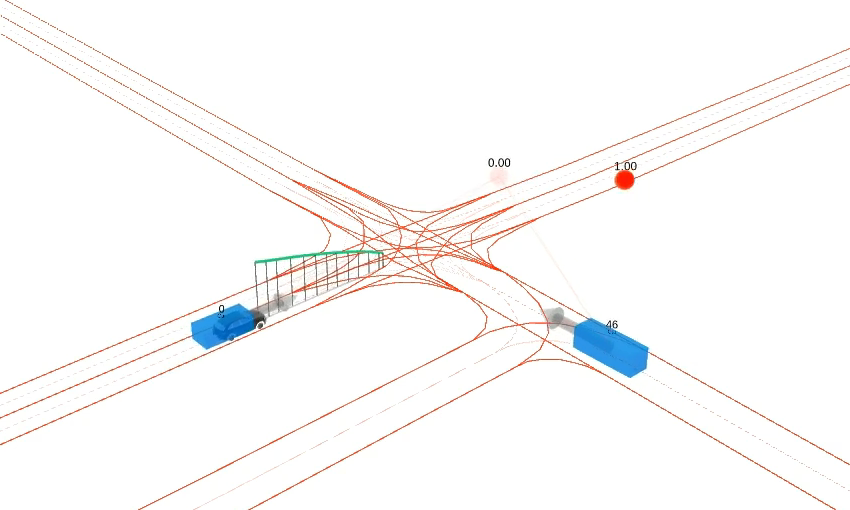}}
        \vspace{-1mm}
        \caption*{Frame 8}
        \vspace{1mm}
        \label{fig:evaluation_uncompliant_demo_ros_h}
    \end{subfigure}

    \begin{subfigure}[b!]{0.24\textwidth}
        \centering
        \frame{\includegraphics[scale=0.137]{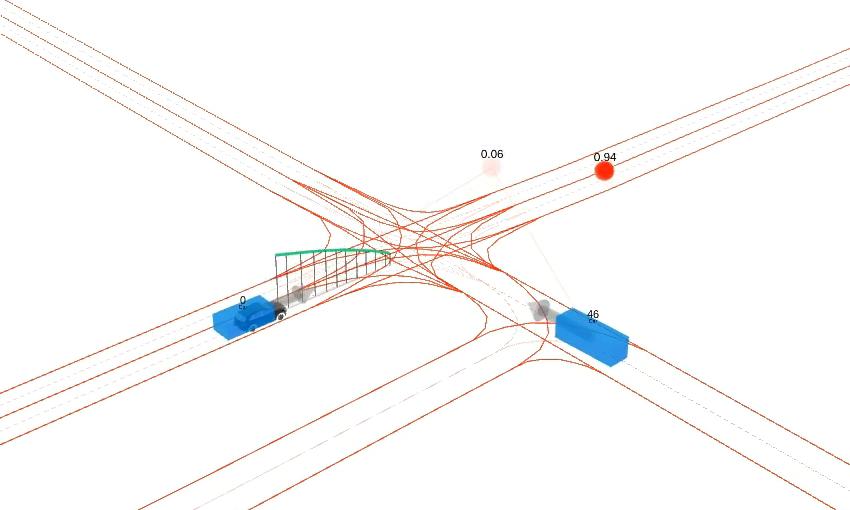}}
        \vspace{-1mm}
        \caption*{Frame 9}
        \vspace{1mm}
        \label{fig:evaluation_uncompliant_demo_ros_i}
    \end{subfigure} \hfill
    \begin{subfigure}[b!]{0.24\textwidth}
        \centering
        \frame{\includegraphics[scale=0.137]{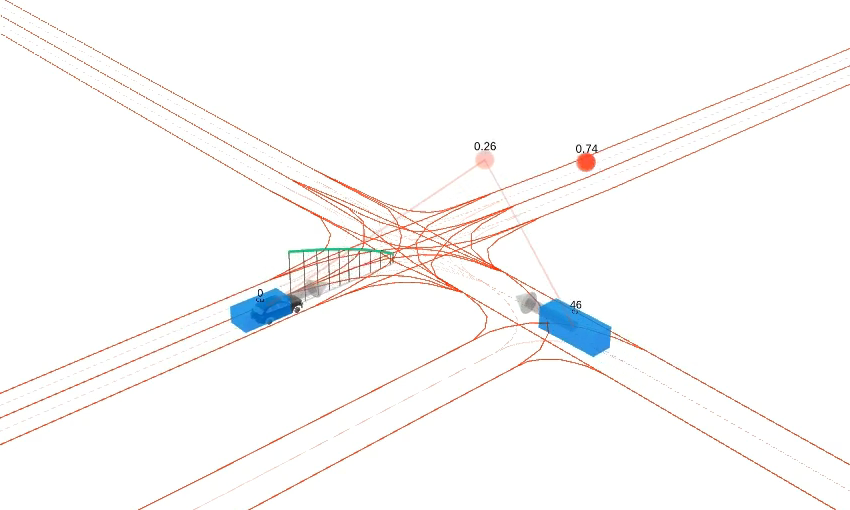}}
        \vspace{-1mm}
        \caption*{Frame 10}
        \label{fig:evaluation_uncompliant_demo_ros_j}
    \end{subfigure} \hfill
    \begin{subfigure}[b!]{0.24\textwidth}
        \centering
        \frame{\includegraphics[scale=0.137]{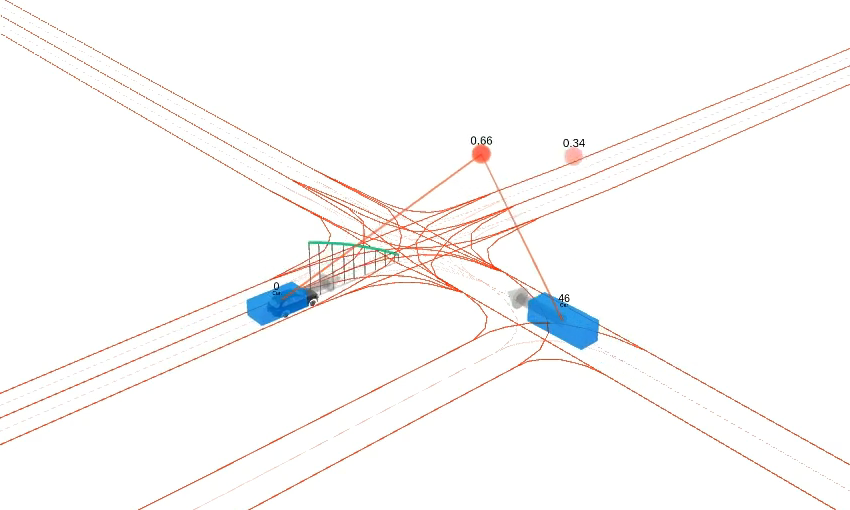}}
        \vspace{-1mm}
        \caption*{Frame 11}
        \label{fig:evaluation_uncompliant_demo_ros_k}
    \end{subfigure} \hfill
    \begin{subfigure}[b!]{0.24\textwidth}
        \centering
        \frame{\includegraphics[scale=0.137]{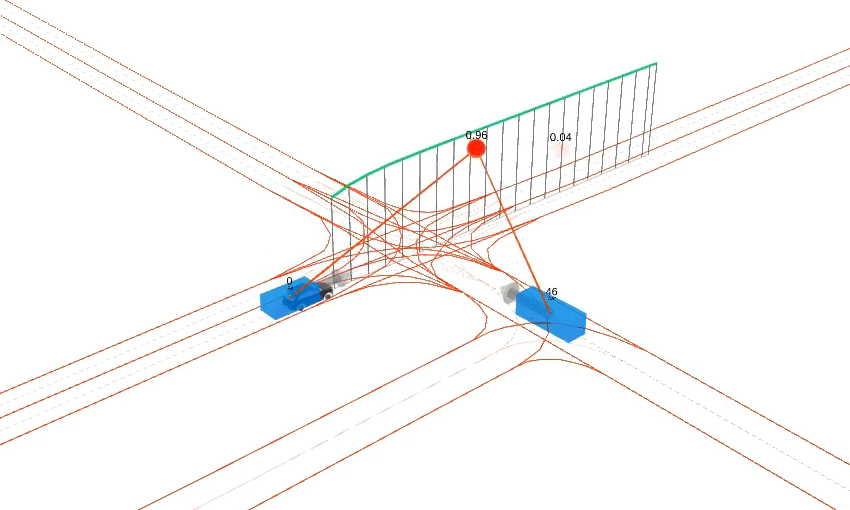}}
        \vspace{-1mm}
        \caption*{Frame 12}
        \label{fig:evaluation_uncompliant_demo_ros_l}
    \end{subfigure}

    \caption[Planned motion during the intersection crossing scenario in RViz]{
        Planned motion during the intersection crossing scenario in RViz.{\footnotemark[2]}
        Frames 3 to 11 illustrate the vehicle's deceleration, enabling the vehicle to come to a full stop before reaching the intersection.
        The probability of the oncoming vehicle crossing through the intersection is displayed at the top of the red circle.
        Frames are incremented every $\SI{0.25}{\second}$, but their numbering is not synchronized with the camera image frames from the previous figure.
    }
    \label{fig:evaluation_uncompliant_demo_ros}
\end{figure*}

\Cref{fig:evaluation_uncompliant_demo_camera} presents a sequence of frames extracted from a video of one test run, illustrating the consistent success observed across all experiments.
In these experiments, the autonomous vehicle uses the probability of the other vehicle crossing the intersection to determine preference weights for its maneuver options.
The planned motion is depicted in \cref{fig:evaluation_uncompliant_demo_ros}.
As the autonomous vehicle approaches the intersection, it recognizes that the oncoming vehicle will not be able to stop comfortably to yield.
Consequently, it maintains its stop constraint, as defined in \cref{eq:chance_constraints_braking_MP_general}.
The weighted maneuver preferences cause the vehicle to reduce its speed during this approach.
Once it becomes evident that the oncoming vehicle will yield, the autonomous vehicle assigns zero weight to its yield maneuver option and releases the stop constraint.

\subsection{Interaction-aware Intersection Crossing}

\begin{figure*}[ht!]
    \vspace{2mm}
    \centering
    \fontsize{7}{8}\selectfont %
    \def\svgwidth{0.95\linewidth}
    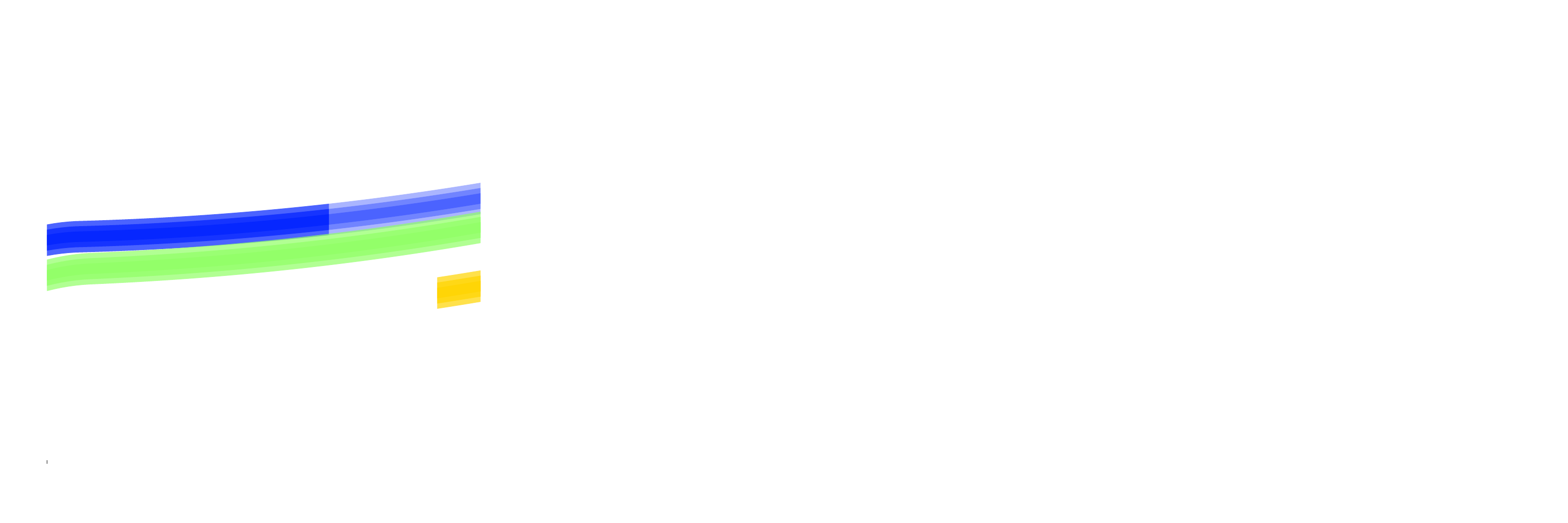
    \caption[An intersection crossing scenario with multiple vehicles.]{
        An intersection crossing scenario with multiple vehicles.
        The ego vehicle is depicted in black, while the vehicles in the environment model $\objects_{\mkern-1.5mu\environment}$ are shown with their position uncertainty and have a semi-transparent rectangle centered at their mean position.
        Vehicles in the scene model $\objects_{\mkern-1.5mu\scene}$ are depicted with opaque rectangles.
        Despite visibility constraints, uncertainty in perception and prediction, the decision-theoretic MPC plans smooth motion profiles and crosses the roundabout.{\footnotemark[3]}
    }
    \label{fig:eval_decision_theoretic_planning_p3iv}
\end{figure*}

The decision-theoretic MPC's cost function facilitates interactive behavior and smooth maneuver transitions, enabling it to adapt to changing predictions and driving goals.
We evaluate these characteristics, along with other features such as planning with a limited field-of-view, using the free and open-source simulation environment P3IV \cite{p3iv}\footnote{Available at {\url{https://github.com/fzi-forschungszentrum-informatik/P3IV}}}.
\Cref{fig:eval_decision_theoretic_planning_p3iv} illustrates a planning instance from a roundabout scenario \cite{zhan2019interaction}.
The vehicle evaluates the probabilities of distinct maneuver options of other vehicles with a particle filter-based prediction module and plans a trajectory accordingly.
The figure consists of three subplots: 1) most likely predictions of vehicles in the scene model together with the planned motion on a path-time diagram, 2) vehicle positions and uncertainties at the time of planning, and 3) the profile of the planned motion.
These results showcase consistently smooth motion profiles.

\clearpage

\begin{textblock*}{\textwidth}(16.6mm,\dimexpr\paperheight-18.7mm\relax)
    \footnotesize
    \textsuperscript{3}Video available online at {\url{https://youtu.be/lOpE5uldJiI}}
\end{textblock*}

\FloatBarrier \newpage
\clearpage

\footnotetext[4]{Video available online at {\url{https://youtu.be/iqZvPMBpUZQ}}}

\subsection{Comparison of Vehicle Models for Replanning}

One of the major challenges in developing planning algorithms through numerical optimization is the modeling and initialization of the problem.
The underlying vehicle model, in addition to the modeled objective function and constraints, significantly influences the solution.
Therefore, we evaluate the runtimes of various vehicle models during replanning for different initializations.
\Cref{tab:cpu_times} presents results using an Intel i7-4710MQ CPU.
While it is intuitive that more complex models increase runtimes, it is surprising to discover that under certain conditions, even the linear kinematic vehicle model can exhibit poor convergence and significantly longer runtimes.
This often occurs when the model's initialization -- and thus, its linearization -- is far from the optimum.
Given the necessity to incorporate new environment information during replanning, linearization within receding horizon planning can be a delicate task.
We provide a qualitative comparison of the kinematic vehicle model and dynamic vehicle model in \cref{fig:vehicle_model_scenarios} and \cref{fig:vehicle_model_scenario_results} in the appendix.
In addition to comparing vehicle models, we also examine the difference in runtimes between classic Runge-Kutta and Euler's method as integration schemes.
Our evaluations reveal that the differences in runtimes are negligible.

\begin{table}[h]
    \centering
    \begin{tabular}{lcccccc} %
        \toprule
        \multirow{2}{*}[-1mm]{\textbf{Model}} & \multicolumn{2}{c}{S-Shaped (free)} & \multicolumn{2}{c}{S-Shaped} & \multicolumn{2}{c}{180\textdegree - Corner}                            \\
        \cmidrule(lr){2-3} \cmidrule(lr){4-5} \cmidrule(lr){6-7}
                                              & Avg.                                & Max                          & Avg.                                        & Max    & Avg.   & Max    \\
        \midrule
        Point Mass                            & 21.25                               & 23.16                        & 50.78                                       & 75.21  & 57.68  & 91.86  \\
        Kinematic                             & 26.73                               & 39.95                        & 53.38                                       & 91.76  & 56.82  & 83.66  \\
        Linear Kin.\                          & 27.08                               & 33.55                        & 67.37                                       & 759.09 & 69.19  & 183.52 \\
        Dynamic                               & 57.91                               & 85.95                        & 115.84                                      & 194.49 & 123.24 & 184.21 \\
        \bottomrule
    \end{tabular}

    \caption{
        Comparison of solver times in \SI{}{\milli\second} for various vehicle models across different scenarios during replanning.
        The scenarios are depicted in \cref{fig:vehicle_model_scenarios} and \cref{fig:vehicle_model_scenario_results} in the appendix.
    }
    \label{tab:cpu_times}
\end{table}

\subsection{Solver Algorithms and Runtimes}

Nonlinear optimization algorithms use linear solvers to solve linearized subproblems.
In our runtime analysis, the linear solvers MUMPS \cite{MUMPS}, MA27, MA57 \cite{hsl} performed comparably.
Further qualitative runtime evaluations showed that merely \SIrange[range-phrase = --]{10}{15}{\percent} of the solution time was spent on calculating derivatives.
Despite some works reporting a superior performance of CppAD over CasADi, our experiments did not reveal any significant difference in runtimes.

The computation of the exact Hessian in each iteration carries an additional computational overhead.
Quasi-Newton methods approximate the Hessian, thereby reducing computational cost and memory requirements at the cost of some accuracy and convergence speed.
Our evaluations demonstrated marginally better runtimes favoring full Newton methods.

\section{Discussions}
\label{sec:discussions}

In this section we discuss our design choices and compare them with other options.
We employ full braking maneuver as a fallback plan for collision avoidance.
While swerving is a potential alternative to full braking, under uncertainty it can pose two significant problems:
Firstly, vehicle models are often linearized for Gaussian error propagation and uncertainty modeling.
This could lead to substantial modeling errors when executing maneuvers at the vehicle's handling limits.
Secondly, the area used during swerving might overlap with the occupancy of objects located beyond the visible field.

In motion planning for autonomous driving, certain works use predicted motion profiles of objects to narrow the drivable corridor and address collision avoidance \cite{ziegler2014trajectory}.
Although this approach potentially allows for faster solver times, it hinders the motion planner from processing the environment probabilistically, thereby limiting its interaction capabilities.
Therefore, we did not adopt this approach in our work.

During our development process, we also analyzed the use of a continuously differentiable distance functions \cite{ziegler2014trajectory} to directly calculate longitudinal positions from Cartesian coordinates, as defined in \cref{eq:cartesian2frenet}.
While this approach initially seemed promising, especially in problems with conflicting objectives, it occasionally led to poor convergence, resulting in unstable behavior.

In receding horizon planning, the vehicle's environment is constantly evolving, leading to changes in constraint values.
This complicates the use of previous solution for initialization, especially if constraints are applied for collision avoidance over the entire planning horizon, potentially rendering it infeasible.
This problem can be addressed in two alternative ways.
The first option is to initialize the problem with a profile that guarantees feasibility, such as an emergency plan.
However, such a \textit{cold} start increases computational cost in receding horizon planning.
Another option is to initialize the free parameters with a simple driving model like IDM.
This approach, though, limits reactive capabilities to the longitudinal direction.
To address these limitations, we initialize with the solution of the previous timestep and apply fallback maneuver constraints only to the immediate time horizon.
Additionally, we use IPMs for solving the problem, which are tolerant of infeasible initialization.

In our evaluations, we demonstrated results for scenarios with multiple maneuver options.
In scenarios with a single maneuver option, the decision-theoretic planner assigns zero weight to the second maneuver, making its parameters ineffective.
It is important to note that the weights in our multi-objective optimization problem are scenario dependent.
While conventional MPC approaches can achieve similar results with careful parameterization, our formulation's core strength lies in its ability to find a unified solution for multiple maneuvers.

Finally, it is crucial to emphasize that the replanning time $\computationTime$ directly affects the ability to reach the desired speed in receding horizon planning under safety constraints.
Ensuring a feasible emergency plan within $\computationTime$ can prevent reaching desired travel speeds, especially when the replanning time is long and the field-of-view is limited.

\section{Conclusion}
\label{sec:conclusion}

Uncertainties due to impaired perception and partial observability often lead to defensive plans.
In this work, we developed a planning approach that eliminates the need to determine the best homotopy in advance.
Instead, it incorporates maneuver preferences in planning.
In this way, it couples decision making with local continuous optimization.

In the context of autonomous driving, our approach can produce maneuver-neutral motion that serves as an intermediate choice between distinct maneuver alternatives.
We have paired this approach with inevitable collision states to develop a proactive, chance-constrained safe motion planning formulation.
Consequently, it offsets the limitations of perception and prediction while maintaining interaction-aware driving and safety.
Evaluations from both driving experiments and simulations derived from real-world driving scenarios demonstrate that the presented approach proves particularly beneficial whenever the current environment information entails significant levels of uncertainty.

In this work, our focus was primarily on urban driving scenarios, though the formulations presented can be seamlessly extended to highway lane-change situations.
While our evaluations centered around traffic participants that are vehicles, the approach remains adaptable to other types, such as cyclists and pedestrians.
One limitation of the present work is that we restricted ourselves to only two maneuver alternatives.
The extension to more than two maneuvers involves additional steps.
Firstly, environment modeling and the identification of non-conflicting maneuvers can become complex.
Secondly, as the computational complexity increases with the number of optimization parameters, maintaining a sufficient planning rate might become challenging.
Future work could use a learning-based approach to identify maneuver preferences and compare the resulting planner with end-to-end planning approaches. %

\section*{Acknowledgments}

We thank Ole Salscheider for setting up the vehicle-to-vehicle communication and Florian Kuhnt for providing motion predictions for the vehicle experiments.
We further thank Martin Lauer, Mykel J.\ Kochenderfer and anonymous reviewers for their comprehensive review.

\clearpage
\onecolumn
\appendix
\label{sec:appendix}

\begin{figure*}[h!]
    \centering
    \begin{subfigure}[t]{0.3\textwidth}
        \centering
        \input{1_figures/postpone_scenarios/s_shape_0200.tex}
    \end{subfigure} %
    \begin{subfigure}[t]{0.3\textwidth}
        \centering
        \input{1_figures/postpone_scenarios/s_shape_1200.tex}
    \end{subfigure}%
    \begin{subfigure}[t]{0.3\textwidth}
        \centering
        \input{1_figures/postpone_scenarios/s_shape_2200.tex}
    \end{subfigure}
    \caption{
        Simulation results in an S-shaped corridor with obstacles of uncertain existence.
        The obstacles are divided into two groups, but only one group actually exists in reality.
        The vehicle is unaware of which group is real, and thus, treats the corresponding maneuver options with equal preference.
        The color of each maneuver option match the color of obstacles considered during planning.
        To ensure safety and prevent potential collisions, all obstacles are accounted for in the time interval $[0, \Time_{2\text{pin}}]$.
        For this reason, the corresponding trajectories maintain a sufficient distance from the obstacles.
        Unlike conventional MPCs, the first few meters of the trajectories are optimized for both maneuver options.
        After this neutral part, the trajectories diverge.
    }
    \label{fig:homotopy_free_s}
\end{figure*}

\begin{figure*}[h!]
    \centering
    \begin{subfigure}[t]{0.3\textwidth}
        \centering
        \input{1_figures/postpone_scenarios/cornering_180_0200.tex}
    \end{subfigure} %
    \begin{subfigure}[t]{0.3\textwidth}
        \centering
        \input{1_figures/postpone_scenarios/cornering_180_1200.tex}
    \end{subfigure}%
    \begin{subfigure}[t]{0.3\textwidth}
        \centering
        \input{1_figures/postpone_scenarios/cornering_180_2200.tex}
    \end{subfigure}
    \caption{
        Simulation results for collision avoidance with two obstacles during a 180\textdegree-cornering.
        The motion profiles exhibit similar features to those in the previous figure.
    }
    \label{fig:homotopy_free_cornering}
\end{figure*}

\begin{figure*}[h!]
    \centering
    \begin{subfigure}[t]{0.3\textwidth}
        \centering
        \input{1_figures/vehicle_model/s.tex}
    \end{subfigure}%
    \begin{subfigure}[t]{0.3\textwidth}
        \centering
        \input{1_figures/vehicle_model/so.tex}
    \end{subfigure}
    \begin{subfigure}[t]{0.3\textwidth}
        \centering
        \input{1_figures/vehicle_model/co.tex}
    \end{subfigure} %
    \caption{
        Scenarios for vehicle model simulations.
        Motion profiles for a dynamic bicycle model incorporating a linear tire model are compared with those from  kinematic vehicle model.
        The first scenario is an S-shaped curve free of obstacles, the second scenario is the same but with obstacles, and the third scenario is 180\textdegree-cornering with obstacles.}
    \label{fig:vehicle_model_scenarios}
\end{figure*}

\begin{figure*}[h!]
    \centering
    \begin{subfigure}[t]{0.3\textwidth}
        \centering
        \input{1_figures/vehicle_model/s_detail.tex}
    \end{subfigure} %
    \begin{subfigure}[t]{0.3\textwidth}
        \centering
        \input{1_figures/vehicle_model/so_detail.tex}
    \end{subfigure}%
    \begin{subfigure}[t]{0.3\textwidth}
        \centering
        \input{1_figures/vehicle_model/co_detail.tex}
    \end{subfigure}
    \caption{
        Comparison of steering angles $\delta$ for the scenarios tested in the previous figure, presented in the same order.
        The color scheme for the vehicle models remains the same.}
    \label{fig:vehicle_model_scenario_results}
\end{figure*}

\twocolumn
\clearpage
\bibliographystyle{IEEEtran}
\bibliography{3_references/references}
\newpage

\begin{IEEEbiography}[{\includegraphics[width=1in,height=1.25in,clip,keepaspectratio]{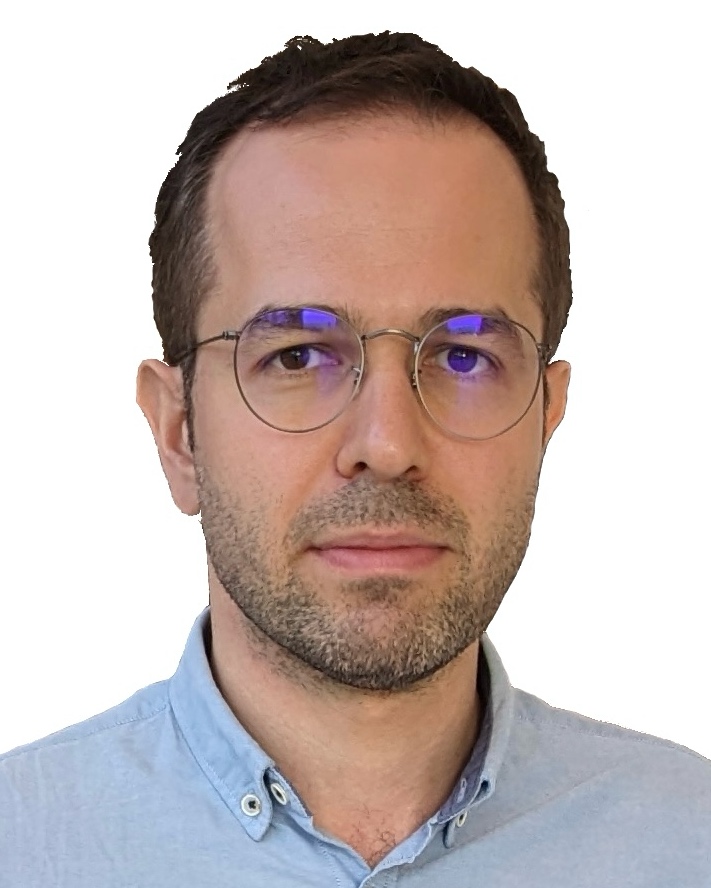}}]{Ömer~Şahin~Taş}
    earned his B.Sc.\ degree from Istanbul Technical University, followed by M.Sc.\ and Ph.D.\ degrees from KIT, with ``very good'' and ``summa cum laude'' distinctions, respectively.
    His dissertation won the UniDAS Science Award and was an IEEE ITS Society Best Dissertation Award finalist.
    After his Ph.D., he held a visiting researcher position at the University of Toronto.
    He has been leading the Mobile Perception Systems department at FZI Forschungszentrum Informatik since 2017 and serving as an Associate Editor for the IEEE Intelligent Vehicles Symposium since 2022.
\end{IEEEbiography}

\begin{IEEEbiography}[{\includegraphics[width=1in,height=1.25in,clip,keepaspectratio]{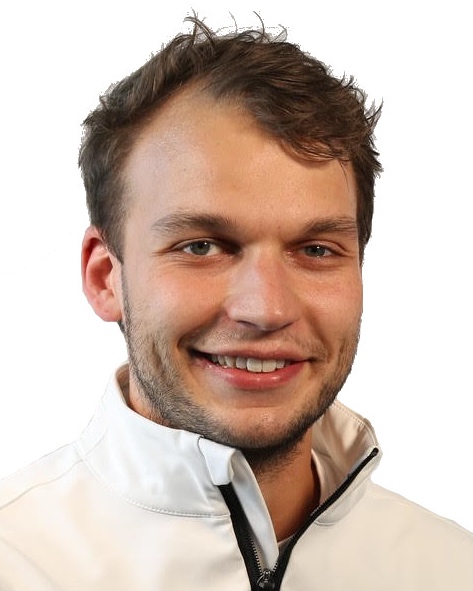}}]{Philipp~Heinrich~Brusius}
    received the M.Sc.\ degree in mechanical engineering from KIT in 2021.
    He is currently working for Porsche Motorsport of Dr.\ Ing.\ h.c.\ F.\ Porsche AG, as part of the Vehicle Science group.
    His work includes the development of simulation methods for the vehicle dynamics analysis.
\end{IEEEbiography}

\begin{IEEEbiography}[{\includegraphics[width=1in,height=1.25in,clip,keepaspectratio]{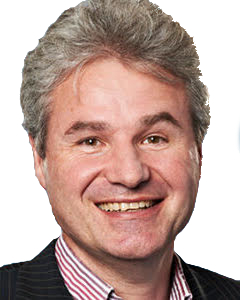}}]{Christoph~Stiller}
    studied electrical engineering toward Diploma degree in Aachen, Germany, and Trondheim, Norway.
    He received the Dr.-Ing. (Ph.D.) from RWTH Aachen University in 1994 and spent a postdoc year at INRS
    in Montreal, Canada. In 1995, he joined the Corporate Research and Advanced Development of Robert Bosch GmbH, Hildesheim, Germany. In 2001, he became Chaired Professor at KIT, Germany.
    He was invited several prestigious universities for extended periods.
    In 2010, he spent three months by invitation at CSIRO in Brisbane, Australia, followed by a four-month sabbatical in 2015 with Bosch RTC and Stanford University in California.
    In 2023, he spent on a six-month sabbatical at the University of California, Berkeley.
    He served as the President of the IEEE Intelligent Transportation Systems Society (2012–2013) and was Vice President before since 2006. He served as the Editor-in-Chief of the IEEE Intelligent Transportation Systems Magazine (2009–2011) and as Associate Editor for the IEEE Transactions on Image Processing (1999–2003), for the IEEE Transactions on Intelligent Transportation Systems (2004-ongoing), and for the IEEE Intelligent Transportation Systems Magazine (2012-ongoing).
    His Autonomous Vehicle AnnieWAY was a finalist in the Urban
    Challenge 2007 and the winner and second winner of the Grand Cooperative
    Driving Challenge 2011 and 2016, respectively.
    In 2013, he collaborated with Daimler on the automated Bertha Benz Memorial Tour.
\end{IEEEbiography}
\end{document}